\documentclass{article}

\usepackage{microtype}
\usepackage{graphicx}
\usepackage{subfigure}
\usepackage{booktabs}

\usepackage{hyperref}

\usepackage{amsmath,amsfonts,bm}

\def\1{\bm{1}}

\def\rb{{\textnormal{b}}}

\def\vb{{\bm{b}}}

\DeclareMathAlphabet{\mathsfit}{\encodingdefault}{\sfdefault}{m}{sl}
\SetMathAlphabet{\mathsfit}{bold}{\encodingdefault}{\sfdefault}{bx}{n}

\DeclareMathOperator*{\argmin}{arg\,min}

\let\ab\allowbreak

\usepackage{url}
\numberwithin{equation}{section}
\usepackage{lineno}

\usepackage[utf8]{inputenc} % allow utf-8 input
\usepackage[T1]{fontenc}    % use 8-bit T1 fonts
\usepackage{hyperref}       % hyperlinks
\usepackage{url}            % simple URL typesetting
\usepackage{booktabs}       % professional-quality tables
\usepackage{amsfonts}       % blackboard math symbols
\usepackage{nicefrac}       % compact symbols for 1/2, etc.
\usepackage{microtype}      % microtypography
\usepackage{mathtools}% Loads amsmath
\usepackage{graphicx}
\usepackage{float}
\usepackage{xcolor}
\usepackage{amssymb,amsmath,amsthm,dsfont,bm}
\usepackage[sort, numbers]{natbib}
\usepackage{enumitem}
\usepackage[compact]{titlesec}
\usepackage{sidecap}

\usepackage{color}
\usepackage{tikz}
 
\usepackage{animate}
\usetikzlibrary{arrows.meta}
\usepackage{ifthen}

\usepackage{pgf}
\usepackage{pgffor}
\usepgfmodule{shapes}
\usepgfmodule{plot}
\usetikzlibrary{decorations}
\usetikzlibrary{arrows}
\numberwithin{equation}{section}

\definecolor{mahogany}{rgb}{1,0.5,0}
\definecolor{darkgreen}{rgb}{0.2,0.5,0.2}
\newcommand\numberthis{\addtocounter{equation}{1}\tag{\theequation}}

\def \RR{\mathbb{R}}
\def \PP{\mathbb{P}}
\def \EE{\mathbb{E}}

\renewcommand{\ab}{\mathbf{a}}
\newcommand{\bbb}{\mathbf{b}}

\newcommand{\fb}{\mathbf{f}}
\newcommand{\gb}{\mathbf{g}}

\newcommand{\pb}{\mathbf{p}}
\newcommand{\qb}{\mathbf{q}}
\renewcommand{\rb}{\mathbf{r}}

\newcommand{\ub}{\mathbf{u}}
\renewcommand{\vb}{\mathbf{v}}
\newcommand{\wb}{\mathbf{w}}
\newcommand{\xb}{\mathbf{x}}
\newcommand{\yb}{\mathbf{y}}
\newcommand{\zb}{\mathbf{z}}

\newcommand{\Ab}{\mathbf{A}}
\newcommand{\Bb}{\mathbf{B}}

\newcommand{\Gb}{\mathbf{G}}
\newcommand{\Hb}{\mathbf{H}}
\newcommand{\Ib}{\mathbf{I}}
\newcommand{\Jb}{\mathbf{J}}
\newcommand{\Kb}{\mathbf{K}}

\newcommand{\Nb}{\mathbf{N}}

\newcommand{\Sbb}{\mathbf{S}}

\newcommand{\Vb}{\mathbf{V}}
\newcommand{\Wb}{\mathbf{W}}

\newcommand{\bLambda}{\bm{\Lambda}}

\newcommand{\cB}{\mathcal{B}}

\newcommand{\cV}{\mathcal{V}}

\newcommand{\pdv}[2]{\frac{\partial #1}{ \partial #2}}

\renewcommand{\argmin}{\mathop{\mathrm{argmin}}}

\def \var {\mathrm{Var}}
\def \ind {\mathds{1}}

\newenvironment{proofLemma}[1]{\noindent{\bf Proof of Lemma #1: \\}}{\qed\bigskip}
\newenvironment{proofTheorem}[1]{\noindent{\bf Proof of Theorem #1:\\}}{\qed\bigskip}
\newenvironment{proofObs}[1]{\noindent{\bf Proof of Observation #1:\\}}{\qed\bigskip}

\newenvironment{proofProp}[1]{\noindent{\bf Proof of Proposition #1: \\}}{\qed\bigskip}

\newtheorem{theorem}{Theorem}
\newtheorem{proposition}{Proposition}

\newtheorem{property}{Property}

\newtheorem*{remark*}{Remark}

\newtheorem{lemma}{Lemma}

\newtheorem{assumption}{Assumption}
\newtheorem{observation}{Observation}

\numberwithin{theorem}{section}
\numberwithin{lemma}{section}

\def\figscale{0.28}

\newcommand{\gradflow}{
\begin{tikzpicture}[scale=\figscale, every node/.style={scale=1}]
    \def\a{-20}
    \def\b{20}
    \def\c{-20}
   \draw[->,>=latex, red,line width = 0.5mm,loosely dashed,domain=\a:\b] plot ({8*cos(\x)}, {8*sin(\x)});
      \draw[->,>=latex, dashed, line width= 0.3mm] (0,0) -- ({8*cos(\a)}, {8*sin(\a)}) node[below= 4pt]{$\vb_k(0)$} ;
       \draw[->,>=latex, line width= 0.3mm] ({8*cos(\c)}, {8*sin(\c)}) -- ({8*cos(\c) -4*sin(\c)}, {8*sin(\c)+4*cos(\c)}) node[right]{$\frac{d\vb_k}{dt}(0)$} ;
       \draw[->,>=latex, thick, xshift=6cm,yshift=-1.5cm] (120:1cm) arc (\a:\b:2) ; node[below] ;
        \node at (0,0) [circle,fill,inner sep=1.5pt]{};
        \draw[->,>=latex, dashed, line width= 0.3mm] (0,0) -- ({8*cos(\b)}, {8*sin(\b)}) node[above=5pt]{$\vb_k(t)$} ;
        
    \filldraw[fill=green!20] (0,0) -- ({1*cos(\a)}, {1*sin(\a)}) arc (\a:\b:1) -- cycle node[left=4pt] {$\alpha$};
      \filldraw[fill=blue!20, opacity= 0.2] (0,0) -- ({8*cos(\a)}, {8*sin(\a)}) arc (\a:\b:8) -- cycle node[left=4 pt] {$\alpha$};

\end{tikzpicture}
}
\newcommand{\discretegrad}{
\begin{tikzpicture}[scale=\figscale, every node/.style={scale=1}]
\coordinate (O) at (0,0);
\coordinate (A) at (4,0);
\coordinate (B) at (4,4);
\coordinate (C) at (1.17,6.828);
\coordinate (D) at (-2.7,7.4);

\node at (0,0) [circle,fill,inner sep=1.5pt]{};
\draw[->,,>=latex][loosely dashed] (O)--(A) node[below= 4pt ]{$\vb_k(0)$};
\draw[->,,>=latex ][loosely dashed] (O)--(B);
\draw[->,,>=latex ][loosely dashed] (O)--(C);
\draw[->,>=latex ][loosely dashed] (O)--(D);
\draw[->,>=latex ][line width= 0.5mm] (A)--(B);
\draw[->,>=latex ][line width= 0.5mm]  (B)--(C) node[below = 10pt, right = 15pt ]{$-\nabla _{\vb_k}L$};

\draw[->,>=latex ][line width= 0.5mm]  (C)--(D) node[above= 4pt ]{$\vb_k(s)$};

\draw [fill=blue!20, opacity= 0.2] (O)--(A)--(B)--(C)--(D);

 \draw[->,>=latex, thick, xshift=1.5cm,yshift=2cm] (120:1cm) arc (0:80:1) node[below]{};
\end{tikzpicture}
}

\usepackage[accepted]{icml2020}
\icmltitlerunning{Optimization Theory for ReLU Neural Networks Trained with Normalization Layers}

\begin{document}

\twocolumn[
\icmltitle{Optimization Theory for ReLU Neural Networks\\ Trained with Normalization Layers}

\icmlsetsymbol{equal}{*}

\begin{icmlauthorlist}
\icmlauthor{Yonatan Dukler}{to}
\icmlauthor{Quanquan Gu}{cs}
\icmlauthor{Guido Mont\'ufar}{to,stat,mpi}

\end{icmlauthorlist}

\icmlaffiliation{to}{Department of Mathematics, UCLA, Los Angeles, CA 90095.}

\icmlaffiliation{cs}{Department of Computer Science, UCLA, Los Angeles, CA 90095.}

\icmlaffiliation{stat}{Department of Statistics, UCLA, Los Angeles, CA 90095.}

\icmlaffiliation{mpi}{Max Planck Institute for Mathematics in the Sciences, 04103 Leipzig, Germany}

\icmlcorrespondingauthor{Yonatan Dukler}{ydukler@math.ucla.edu} 

\icmlcorrespondingauthor{Quanquan Gu}{qgu@cs.ucla.edu} 

\icmlcorrespondingauthor{Guido Mont\'ufar}{montufar@math.ucla.edu}

\icmlkeywords{Machine Learning, ICML}

\vskip 0.3in
]

\printAffiliationsAndNotice{}

\begin{abstract}
The success of deep neural networks is in part due to the use of normalization layers. Normalization layers like Batch Normalization, Layer Normalization and Weight Normalization are ubiquitous in practice, as they improve generalization performance and speed up training significantly. Nonetheless, the vast majority of current deep learning theory and non-convex optimization literature focuses on the un-normalized setting, where the functions under consideration do not exhibit the properties of commonly normalized neural networks. In this paper, we bridge this gap by giving the first global convergence result for two-layer neural networks with ReLU activations trained with a normalization layer, namely Weight Normalization. Our analysis shows how the introduction of normalization layers changes the optimization landscape and can enable faster convergence as compared with un-normalized neural networks.  
\end{abstract}

\section{Introduction}
Dynamic normalization in the training of neural networks amounts to the application of an intermediate normalization procedure between layers of the network. Such methods have become ubiquitous in the training of neural nets since in practice they significantly improve the convergence speed and stability. 
This type of approach was popularized with the introduction of Batch Normalization (BN) \cite{ioffe2015batch} which implements a dynamic re-parametrization normalizing the first two moments of the outputs at each layer over mini-batches. A plethora of additional normalization methods followed BN, notably including Layer Normalization (LN) \cite{ba2016layer} and Weight Normalization (WN) \cite{salimans2016weight}. 
Despite the impressive empirical results and massive popularity of dynamic normalization methods, explaining their utility and proving that they converge when training with non-smooth, non-convex loss functions has remained an unsolved problem. In this paper we provide sufficient conditions on the data, initialization, and over-parametrization for dynamically normalized ReLU networks to converge to a global minimum of the loss function. For the theory we present we focus on WN, which is a widely used normalization layer in training of neural networks. 
WN was proposed as a method that emulates BN. It normalizes the input weight vector of each unit and separates the scale into an independent parameter. 
The WN re-parametrization is very similar to BN (see Section~\ref{section:weight_normalization_dynamics}) and benefits from similar stability and convergence properties. Moreover, WN has the advantage of not requiring a batch setting, therefore considerably reducing the computational overhead that is imposed by BN \cite{gitman2017comparison}.

When introducing normalization methods, the function parametrization defined by the network becomes scale invariant in the sense that re-scaling of the weights does not change the represented function. 
This re-scaling invariance changes the geometry of the optimization landscape drastically. To better understand this we analyze weight normalization in a given layer. 

We consider the class of 2-layer ReLU neural networks which represent functions $f\colon \RR^{d} \rightarrow \RR$ parameterized by $(\Wb, \mathbf{c}) \in \RR^{m \times d} \times \RR^{m}$ as %of the form 
\begin{align}\label{eq:un-norm-net-class}
f (\xb; \Wb,\mathbf{c}) = \frac{1}{\sqrt{m}} \sum_{k =1}^{m} c_k \sigma(\wb^{\top}_{k}\xb). 
\end{align}
Here we use the ReLU activation function $\sigma(s) = \max\{s,0\}$ \cite{nair2010rectified}, 
$m$ denotes the width of the hidden layer, 
and the output is normalized accordingly by a factor $\sqrt{m}$. 
We investigate gradient descent training with WN for \eqref{eq:un-norm-net-class}, which re-parametrizes the functions in terms of $(\Vb,\gb, \mathbf{c}) \in \RR^{m \times d} \times \RR^m \times \RR^m$ as 
\begin{align}\label{eqn:netclass}
f (\xb; \Vb,  \gb, \mathbf{c}) = \frac{1}{\sqrt{m}} \sum_{k =1}^{m} c_k  \sigma\bigg(g_k \cdot \frac{\vb^{\top}_{k}\xb}{\|\vb_k\|_2}\bigg). 
\end{align}
This gives a similar parametrization to \cite{du2017convolutional} that study convergence of gradient optimization of convolutional filters on Gaussian data. 
We consider a regression task, the $L^2$ loss, a random parameter initialization, and focus on the over-parametrized regime, meaning that $m>n$, where $n$ is the number of training samples. Further, we make little to no assumptions about the data.  

The neural network function class \eqref{eq:un-norm-net-class} has been studied in many papers including \cite{arora2019fine, du2018gradient,zhang2019fast, wu2019global} along with other similar over-parameterized architectures \cite{allen2018learning, li2018learning,du2017convolutional}. An exuberant series of recent works prove that feed-forward ReLU networks converge to zero training error when trained with gradient descent from random initialization. 
Nonetheless, to the best of our knowledge, there are no proofs that ReLU networks trained with \textit{normalization} on general  data converge to a global minimum. This is in part because normalization methods completely change the optimization landscape during training.
Here we show that neural networks of the form given above converge at linear rate when trained with gradient descent and WN. 
The analysis is based on the over-parametrization of the networks, 
which allows for guaranteed descent while the gradient is non-zero. 

For regression training, a group of papers studied the trajectory of the networks' predictions and showed that they evolve  via a ``neural tangent kernel'' (NTK) as introduced by \citet{jacot2018neural}. The latter paper studies neural network convergence in the continuous limit of infinite width over-parametrization, while the works of \cite{du2018gradient, arora2019fine, wu2019global,zhang2019fast, oymak2019towards} analyze the finite width setting. 
For finite-width over-parameterized networks, the training evolution also exhibits a kernel that takes the form of
a Gram matrix. 
In these works, the convergence rate is dictated by the least eigenvalue of the kernel. 
We build on this fact, and also on the general ideas of the proof of \cite{du2018gradient} and the refined work of \cite{arora2019fine}.

In this work we analyze neural network optimization with weight normalization layers. We rigorously derive the dynamics of weight normalization training and its convergence from the perspective of the neural tangent kernel. 
Compared with un-normalized training, we prove that normalized networks follow a modified kernel evolution that features a ``length-direction'' decomposition of the NTK. This leads to two convergence regimes in WN training and explains the utility of WN from the perspective of the NTK. 
In the settings considered, WN significantly reduces the amount of over-parametrization needed for provable convergence, as compared with un-normalized settings. Further, we present a more careful analysis that leads to improved over-parametrization bounds as compared with \cite{du2018gradient}. 

The main contributions of this work are:
\begin{itemize}[leftmargin=*]
  \item We prove the first general convergence result for 2-layer ReLU networks trained with a normalization layer and gradient descent. Our formulation does not assume the existence of a teacher network and has only very mild assumptions on the training data. 
    \item 
    We hypothesize the utility of normalization methods via a decomposition of the neural tangent kernel. In the analysis we highlight two distinct convergence regimes and show how Weight Normalization can be related to natural gradients and enable faster convergence. 

   \item
   We show that finite-step gradient descent converges for all weight magnitudes at initialization. Further, we significantly reduce the amount of over-parametrization required for provable convergence as compared with un-normalized training. 

\end{itemize}

The paper is organized as follows. In Section~\ref{section:weight_normalization_dynamics} we provide background on WN and derive key evolution dynamics of training in Section~\ref{subsec:evolution_dynamics}. We present and discuss our main results, alongside with the idea of the proof, in Section~\ref{section:main}. We discuss related work in Section~\ref{sec:related_works}, and offer a discussion of our results and analysis in Section~\ref{section:discussion}. 
Proofs are presented in the Appendix. 

\section{Weight Normalization}
\label{section:weight_normalization_dynamics}
Here we give an overview of the WN procedure and review some known properties of normalization methods. 
\paragraph{Notation} 
We use lowercase, lowercase boldface, and uppercase boldface letters to denote scalars, vectors
and matrices respectively. We denote the Rademacher distribution as $U\{1,-1 \}$ and write $N(\bm{\mu}, \mathbf{\Sigma})$ for a Gaussian with mean $\bm{\mu}$ and covariance $\mathbf{\Sigma}$.  Training points are denoted by $\xb_1, \ldots, \xb_n \in \RR^{d}$ and parameters of the first layer by $\vb_k \in \RR^{d}$, $k=1,\ldots, m$. We use $\sigma(x) \coloneqq \max \{x, 0\}$, and write $\|\cdot\|_2, \|\cdot\|_{F}$ for the spectral and Frobenius norms for matrices. $\lambda_{\min}(\Ab)$ is used to denote the minimum eigenvalue of a matrix $\Ab$ and $\langle\cdot,\cdot\rangle$ denotes the Euclidean inner product. For a vector $\vb$ denote the $\ell_2$ vector norm as $\|\vb\|_2$ and for a positive definite matrix $\Sbb$ define the induced vector norm $\|\vb \|_{\Sbb}\coloneqq \sqrt{\vb^\top \Sbb \vb }$. The projections of $\xb$ onto $\ub$ and $\ub^\perp$ are defined as $\xb^{\ub} \coloneqq \frac{\ub \ub^\top\xb}{\|\ub\|_2^2}$, $\xb^{\ub^{\perp}} \coloneqq \big(\Ib - \frac{\ub \ub^\top}{\|\mathbf{u}\|_2^2}\big)\xb$. Denote the indicator function of event $A$ as $\ind_{A}$ and for a weight vector at time $t$, $\vb_k(t)$, and data point $\xb_i$ we denote $\ind_{ik}(t) \coloneqq \ind_{\{\vb_k(t)^\top \xb_i\ge~ 0\}}$. 

\paragraph{WN procedure}
For a single neuron $\sigma(\wb^\top \xb)$, WN re-parametrizes the weight $\wb\in\RR^d$ in terms of $\vb \in \RR^d$, $g\in\RR$ as
\begin{equation}\label{eq:wn}
\wb( \vb ,g) =g \cdot \frac{\vb}{\|\vb\|_2}, \quad \sigma \bigg(g \cdot \frac{\vb^\top \xb}{\|\vb\|_2} \bigg).
\end{equation}

This decouples the magnitude and direction of each weight vector (referred as the ``length-direction'' decomposition). In comparison, for BN each output $\wb^\top \xb$ is normalized according to the average statistics in a batch. We can draw the following analogy between WN and BN if the inputs $\xb_i$ are centered ($\EE \xb = \mathbf{0}$) and the covariance matrix is known ($\EE \xb\xb^\top = \Sbb$). In this case, batch training with BN amounts to
\begin{align}
\sigma\Bigg( \gamma \cdot \frac{\wb^\top \xb}{\sqrt{\EE_{\xb} \big( \wb ^\top \xb \xb ^\top \wb \big)}} \Bigg) 
&= \sigma \bigg( \gamma \cdot \frac{\wb^\top \xb}{ \sqrt{\wb^\top \Sbb \wb} } \bigg) \label{eq:bn}\\
&= \sigma \bigg( \gamma \cdot \frac{\wb^\top \xb}{ \|\wb\|_{\Sbb} } \bigg).\nonumber
\end{align}
From this prospective, WN is a special case of \eqref{eq:bn} with $\Sbb = \Ib$~\cite{salimans2016weight,kohler2018towards}. 

\paragraph{Properties of WN}
We start by giving an overview of known properties of WN that will be used to derive the gradient flow dynamics of WN training.

\begin{figure}[H]
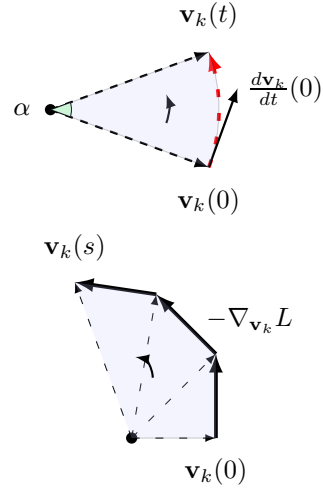

 \centering
    \centering
    \gradflow
 
    \discretegrad
  \caption{WN updates for gradient flow and gradient descent. For gradient flow, the norm of the weights are preserved, i.e., $\|\vb_k(0)\|_2 = \|\vb_k(t)\|_2$ for all $t>0$.
For gradient descent, the norm of the  weights $\|\vb_k(s)\|_2$ is increasing with $s$.
  }
    \label{fig:weight_norm}
\end{figure}

For re-parametrization \eqref{eq:wn} of a network function $f$ that is initially parameterized with a weight $\wb$, 
the gradient $\nabla_\wb f$ relates to the gradients $\nabla_\vb f,~\frac{\partial f}{\partial g}$ by the identities 
\begin{align*}
\nabla_{\vb} f = \frac{g}{\|\vb\|_2}(\nabla_{\wb} f)^{\vb^\perp}, \quad ~ \frac{\partial f}{\partial g} = (\nabla_{\wb} f)^\vb.
\end{align*}
This implies that $\nabla_{\vb} f \cdot \vb = 0$ for each input $\xb$ and  parameter $\vb$. For gradient flow, this orthogonality results in $\|\vb(0)\|_2= \|\vb(t)\|_2$ for all $t$. For gradient descent (with step size $\eta$) the discretization in conjunction with orthogonality leads to increasing parameter magnitudes during training \cite{arora2018theoretical, hoffer2018norm, salimans2016weight}, as illustrated in Figure~\ref{fig:weight_norm}, 
\begin{align}\label{eq:inc}
    \|\vb(s+1)\|^2_2= \|\vb(s)\|^2_2+\eta^2 \|\nabla_{\vb}f\|^2_2\ge \|\vb(s)\|^2_2.
\end{align}

\paragraph{Problem Setup}
We analyze \eqref{eq:un-norm-net-class} with WN training  \eqref{eqn:netclass}, so that 
\begin{align*}
f (\xb; \Vb, \mathbf{c}, \gb) = \frac{1}{\sqrt{m}} \sum_{k =1}^{m} c_k  \sigma\bigg(g_k \cdot \frac{\vb^{\top}_{k}\xb}{\|\vb_k\|_2}\bigg).
\end{align*}
We take an initialization in the spirit of \cite{salimans2016weight}:
\begin{align}
\begin{split}
    \label{eq:init}
   \vb_k(0) \sim N(0,\alpha^2 \Ib),\quad c_k \sim U\{-1,1\}, \\
   \quad \text{and} \quad g_k(0) = \|\vb_k(0)\|_2/\alpha . 
\end{split}
\end{align}
Where $\alpha^2$ is the variance of $\vb_k$ at initialization. The initialization of $g_k(0)$ is therefore taken to be independent of $\alpha$. We remark that the initialization \eqref{eq:init} gives the same initial output distribution as in methods that study the un-normalized network class \eqref{eq:un-norm-net-class}. The parameters of the network are optimized using the training data $\{(\xb_1, y_1), \dots ,(\xb_n,y_n)\}$
with respect to the square loss 
\begin{align}\label{eq:regression}
    L(f) = \frac{1}{2}\sum_{i=1}^{n}(f (\xb_i) - y_i)^2 = \frac{1}{2}\|\fb - \yb \|_2^2,
\end{align}
where $\fb = (f_1,\ldots,f_n)^\top= (f(\xb_1),\ldots,f(\xb_n))^\top$ and $\yb = (y_1,\ldots,y_n)^\top$. 

\section{Evolution Dynamics}
\label{subsec:evolution_dynamics}
We present the gradient flow dynamics of training \eqref{eq:regression} to illuminate the modified dynamics of WN as compared with vanilla gradient descent. In Appendix \ref{sec:finite_step} we tackle gradient descent training with WN where the predictions' evolution vector $\frac{d\fb}{dt}$ is replaced by the finite difference $\fb (s+1) -\fb (s)$.
For gradient flow, each parameter is updated in the negative direction of the partial derivative of the loss with respect to that parameter. The optimization dynamics give
\begin{equation}\label{eq:gradient flow}
\frac{d \vb_k}{dt}
= -\pdv{L}{\vb_k},\quad \frac{d g_k}{dt} = -\pdv{L}{g_k}.
\end{equation}
We consider the case where we fix the top layer parameters $c_k$ during training. In the over-parameterized settings we consider, the dynamics of $c_k$ and $g_k$ turn out to be equivalent.\\
To quantify convergence, we monitor the time derivative of the $i$-th prediction, which is computed via the chain rule as 
\begin{align*}
    \pdv{f_i}{t} = \sum_{k=1}^{m} \pdv{f_i}{\vb_k} \frac{d\vb_k}{dt} +  \pdv{f_i}{g_k} \frac{dg_k}{dt}.
\end{align*}
Substituting \eqref{eq:gradient flow} into the $i$-th prediction evolution and grouping terms yields 
\begin{align}\label{eq:TvTg}
\pdv{f_i}{t} =-\underbrace{\sum_{k=1}^{m} \pdv{f_i}{\vb_k} \pdv{L}{\vb_k}}_{T_{\vb}^{i}} -  \underbrace{\sum_{k=1}^{m}   \pdv{f_i}{g_k} \pdv{L}{g_k}}_{T_{g}^{i}}.
\end{align}
The gradients of $f_i$ and $L$ with respect to $\vb_k$ are written explicitly as
\begin{align*}
\pdv{f_i}{\vb_k}(t)  &= \frac{1}{\sqrt{m}}\frac{c_k \cdot g_k(t)}{\|\vb_k(t)\|_2} \cdot \xb_i^{\vb_k(t)^{\perp}}\ind_{ik}(t), \\
\pdv{L}{\vb_k}(t) &= \frac{1}{\sqrt{m}}\sum_{i= 1}^{n} (f_i(t) -y_i)  \frac{ c_k\cdot g_k(t) }{\|\vb_k(t)\|_2}\xb_i^{\vb_k(t)^{\perp}}\ind_{ik}(t).  
\end{align*}
Defining the $\vb$-orthogonal Gram matrix $\Vb(t)$ as 
\begin{align}
\begin{split}
\label{eq:defineV}
& \Vb_{ij}(t) =  \\
& \frac{1}{m} \sum_{k = 1}^{m} \bigg(\frac{\alpha c_k\cdot g_k(t)}{\|\vb_k(t)\|_2}\bigg)^{2}  \big \langle \xb_i^{\vb_k(t)^{\perp}},~ \xb_j^{\vb_k(t)^{\perp}} \big \rangle
\ind_{ik}(t) \ind_{jk}(t), 
\end{split}
\end{align}
we can compute $T_{\vb}^i$ as 
\begin{align*}
T_{\vb}^i(t) = \sum_{j=1}^{n}\frac{\Vb_{ij}(t)}{\alpha^2} (f_j(t) - y_j). 
\end{align*}
Note that $\Vb(t)$ is the induced neural tangent kernel \cite{jacot2018neural} for the parameters $\vb$ of WN training. While it resembles the Gram matrix $\Hb(t)$ studied in \cite{arora2019fine}, here we obtain a matrix that is not piece-wise constant in $\vb$ since the data points are projected onto the orthogonal component of $\vb$. We compute $T_{\gb}^i$ in \eqref{eq:TvTg} analogously. The associated derivatives with respect to $g_k$ are
\begin{align*}
\pdv{f_i}{g_k}(t) &= \frac{1}{\sqrt{m}}\frac{c_k}{\|\vb_k(t)\|_2}\sigma(\vb_k(t)^\top \xb_i),\\
\quad 
\pdv{L}{g_k}(t) &= \frac{1}{\sqrt{m}}\sum_{j=1}^{n} (f_j(t) - y_j) \frac{c_k}{\|\vb_k(t)\|_2}\sigma(\vb_k(t)^\top \xb_j), 
\end{align*}

and we obtain 
\begin{align*}
T_{\gb}^i(t)  &= \\
\sum_{k =1}^{m}& \frac{1}{m} \sum_{j=1}^{n} \frac{c^2_k(f_j(t) -y_j)}{\|\vb_k(t)\|_2^2}  \sigma(\vb_k(t)^\top \xb_j)\sigma(\vb_k(t)^\top \xb_i). 
\end{align*}
Given that $c_k^2 =1$, define $\Gb(t)$ as
\begin{equation}
\label{eq:defineG}
\Gb_{ij}(t) = \frac{1}{m}\sum_{k =1}^{m}  \frac{  \sigma(\vb_k(t)^\top \xb_i)\sigma(\vb_k(t)^\top \xb_j)
}{\|\vb_k(t)\|_2^2} 
\end{equation}
hence we can write 
\begin{align*}
T_{\gb}^i(t) = \sum_{j =1}^{n} \Gb_{ij}(t)(f_j(t) - y_j). 
\end{align*}
Combining $T_{\vb}$ and $T_{\gb}$, the full evolution dynamics are given by 
\begin{align}\label{eq:flow_evolution}
\frac{d\fb}{dt} = - \bigg(\frac{\Vb(t)}{\alpha^2} + \Gb(t)\bigg) (\fb (t) -\yb).
\end{align}
Denote $\bLambda(t) \coloneqq \frac{\Vb(t)}{\alpha^2} + \Gb(t)$ and write $\frac{d\fb}{dt} = -\bLambda(t) (\fb (t) -\yb)$. 
We note that $\Vb(0), \Gb(0)$, defined in \eqref{eq:defineV}, \eqref{eq:defineG}, are independent of $\alpha$: 
\begin{observation}[$\alpha$ independence]
\label{obs:inv} For initialization \eqref{eq:init} and $\alpha>0$ the Gram matrices $\Vb(0), \Gb(0)$ are independent of $\alpha$. 
\end{observation}
This fact is proved in Appendix~\ref{proof:obs_inv}. 
When training the neural network in \eqref{eq:un-norm-net-class} without WN 
\citep[see][]{du2018gradient,arora2019fine,zhang2019fast}, the corresponding neural tangent kernel $\Hb(t)$ is
defined by $\frac{\partial f_i}{\partial t} = \sum_{k=1}^m \frac{\partial f_i}{\partial \wb_k} \frac{d \wb_k}{dt}=-\sum_{k=1}^m \frac{\partial f_i}{\partial \wb_k} \frac{\partial L}{\partial \wb_k} = -\sum_{j=1}^n \Hb_{ij}(t)(f_j - y_j)$ and takes the form
\begin{align}\label{eq:defineNTK} \Hb_{ij} (t)= \frac{1}{m} \sum_{k=1}^{m} \xb_i^\top \xb_j \ind_{ik}(t) \ind_{jk}(t).
 \end{align}
The analysis presented above shows that vanilla and WN gradient descent are related as follows. 
\begin{proposition}\label{rem:VGH}
Define $\Vb(0)$, $\Gb(0)$, and $\Hb(0)$ as in  \eqref{eq:defineV}, \eqref{eq:defineG}, and \eqref{eq:defineNTK} respectively.
then for all $\alpha > 0$, 
\begin{align*}
\Vb(0) + \Gb(0) = \Hb(0).
\end{align*}
Thus, for $\alpha=1$, 
\begin{align*}
\pdv{\fb}{t} = -\bLambda(0)(\fb (0) - \yb)=  - \Hb(0)(\fb (0) - \yb).
\end{align*}
\end{proposition}
That is, WN decomposes the NTK in each layer into a length and a direction component. We refer to this as the ``length-direction decoupling'' of the NTK, in analogy to \eqref{eq:wn}. 
From the proposition, normalized and un-normalized training kernels initially coincide if $\alpha =1$. We hypothesize that the utility of normalization methods can be attributed to the modified NTK $\bLambda(t)$ that occurs when the WN coefficient, $\alpha$, deviates from $1$. For $\alpha \gg 1$ the kernel $\bLambda(t)$ is dominated by $\Gb(t)$, and for $\alpha \ll 1$  the kernel $\bLambda(t)$ is dominated by $\Vb(t)$. 
We elaborate on the details of this in the next section. In our analysis we will study the two regimes $\alpha > 1$ and $\alpha < 1$ in turn.

\section{Main Convergence Theory}
\label{section:main}
In this section we discuss our convergence theory and main results. From the continuous flow \eqref{eq:flow_evolution}, we observe that the convergence behavior is described by $\Vb(t)$ and  $\Gb(t)$. 
The matrices $\Vb(t)$ and $\Gb(t)$ are positive semi-definite since they can be shown to be covariance matrices. This implies that the least eigenvalue of the evolution matrix $\bLambda(t)= \frac{1}{\alpha^2}\Vb(t) + \Gb(t)$ is bounded below by the least eigenvalue of each kernel matrix,
\begin{align*}
    \lambda_{\min}(\bLambda(t)) \ge \max\{ \lambda_{\min}(\Vb(t)) / \alpha^2, \lambda_{\min}(\Gb(t)) \}.
\end{align*}

For finite-step gradient descent, a discrete analog of evolution \eqref{eq:flow_evolution} holds. However, the discrete case requires additional care in ensuring dominance of the driving gradient terms. For gradient flow, it is relatively easy to see linear convergence is attained by relating the rate of change of the loss to the magnitude of the loss.
Suppose that for all $t \ge 0$,
\begin{align}\label{eq:eigen_condition}
     \lambda_{\min}\big( \bLambda(t) \big) \ge \omega /2,\quad \text{with $\omega > 0$}.
\end{align}
Then the change in the regression loss is written as
\begin{align*}
\frac{d}{dt}\|\fb(t) - \yb\|_2^2 &= 2 (\fb(t) - \yb)^\top  \frac{d \fb (t)}{dt} \\
&= -2(\fb(t) - \yb)^\top \bLambda(t) (\fb(t) - \yb)\\
&\overset{\eqref{eq:eigen_condition}}{\le} -\omega \|\fb(t) - \yb\|_2^2.
\end{align*}
Integrating this time derivative and using the initial conditions yields
\begin{align*}
    \|\fb(t) - \yb\|_2^2 \le \exp(-\omega t) \|\fb(0) -\yb\|_2^2, 
\end{align*}
which gives linear convergence. 
The focus of our proof is therefore showing that \eqref{eq:eigen_condition} holds throughout training. 
\par

By Observation~\ref{obs:inv} we have that $\Vb$ and $\Gb$ are independent 
of the WN coefficient $\alpha$ ($\alpha$ only appears in the $1 /\alpha^2$ scaling of $\bLambda$). This suggests that the kernel $\bLambda(t) = \frac{1}{\alpha^2}\Vb(t) + \Gb(t)$ can be split into two regimes: When $\alpha < 1$ the kernel is dominated by the first term $\frac{1}{\alpha^2}\Vb$, and when $\alpha > 1$ the kernel is dominated by the second term $\Gb$. We divide our convergence result based on these two regimes. 

In each regime, \eqref{eq:eigen_condition} holds if the corresponding dominant kernel, $\Vb(t)$ or $\Gb(t)$, maintains a positive least eigenvalue. Having a least eigenvalue that is bounded from $0$ gives a convex-like property that allows us to prove convergence. 
To ensure that condition \eqref{eq:eigen_condition} is satisfied, for each regime we show that the corresponding dominant kernel is ``anchored'' (remains close) to an auxiliary Gram matrix which we define in the following for $\Vb$ and $\Gb$. 

Define the auxiliary $\vb$-orthogonal and $\vb$-aligned Gram matrices $ \Vb^{\infty}, \Gb^{\infty}$ as 
\begin{align}\label{eq:DefineVinf}
 \Vb^{\infty}_{ij} & \coloneqq  \EE_{\vb \sim N(0,\alpha^2 \Ib)}~ \langle \xb_i^{\vb^\perp}, \xb_j^{\vb^\perp}\rangle \ind_{ik}(0)\ind_{jk}(0), \\
 \Gb^{\infty}_{ij} & \coloneqq \EE_{\vb \sim N(0,\alpha^2 \Ib)}~ \langle \xb_i^{\vb}, \xb_j^{\vb}\rangle \ind_{ik}(0)\ind_{jk}(0).\label{eq:DefineGinf}
\end{align} 
For now, assume that $\Vb^{\infty}$ and $\Gb^{\infty}$ are positive definite with a least eigenvalue bounded below by $\omega$ (we give a proof sketch below). In the convergence proof we will utilize over-parametrization to ensure that $\Vb(t), \Gb(t)$ concentrate to their auxiliary versions so that they are also positive definite with a least eigenvalue that is greater than $\omega/2$. The precise formulations are presented in Lemmas~\ref{lemma:ht} and~\ref{lemma:Gt} that are relegated to Appendix~\ref{appendix:flow_proof}. 

To prove our convergence results we make the assumption that the $\xb_i$s have bounded norm and are not parallel. 
\begin{assumption}[Normalized non-parallel data]
\label{as:parallel}
The data points $(\xb_1, y_1), \ldots, (\xb_n, y_n)$ satisfy $\|\xb_i\|_2\le 1$ and for each index pair $i \ne j$, $\xb_i \ne \beta \cdot \xb_j$~
for all $\beta \in \RR \setminus \{0 \}$. 
\end{assumption}
In order to simplify the presentation of our results, we assume that the input dimension $d$ is not too small, whereby $d\geq 50$ suffices. This is not essential for the proof. Specific details are provided in Appendix~\ref{sec:WN_proofs}. 
\begin{assumption}\label{as:mdelta} 
For data
$\xb_i \in \RR^d$ assume that $d \ge 50$.
\end{assumption}
Both assumptions can be easily satisfied by pre-processing, e.g., normalizing and shifting the data, and adding zero coordinates if needed. 

Given Assumption~\ref{as:parallel}, $\Vb^{\infty}, \Gb^{\infty}$ are shown to be positive definite.
\begin{lemma} \label{lemma:hau}
Fix training data 
$\{(\xb_1, y_1),  \ldots, (\xb_n,y_n) \}$ satisfying Assumption~\ref{as:parallel}.  Then the $\vb$-orthogonal and $\vb$-aligned Gram matrices $\Vb^\infty$ and $\Gb^{\infty}$, defined as in \eqref{eq:DefineVinf} and \eqref{eq:DefineGinf},
are strictly positive definite. We denote the least eigenvalues $\lambda_{\min} (\Vb^{\infty}) \eqqcolon \lambda_0,~\lambda_{\min} (\Gb^{\infty}) \eqqcolon \mu_0$.
\end{lemma}
\paragraph{Proof sketch}
Here we sketch the proof of Lemma~\ref{lemma:hau}. The main idea, is the same as \cite{du2018gradient}, is to regard the auxiliary matrices $\Vb^{\infty}, \Gb^{\infty}$ as the covariance matrices of linearly independent operators. For each data point $\xb_i$, define $\phi_i(\vb) \coloneqq \xb_i^{\vb^\perp}\ind_{\{\xb_i^\top \vb \ge 0\}}$. The Gram matrix $\Vb^{\infty}$ is the covariance matrix of $\{\phi_i\}_{i=1:n}$ taken over $\RR^{d}$ with the measure $N(0, \alpha^2 \Ib)$. Hence showing that $\Vb^{\infty}$ is strictly positive definite is equivalent to showing that $\{\phi_i\}_{i=1, \dots n}$ are linearly independent. Unlike \cite{du2018gradient}, the functionals under consideration are not piecewise constant so a different construction is used to prove independence. Analogously, a new set of operators, $\theta_i(\vb)\coloneqq \sigma (\xb_i^{\vb})$, 
is constructed for  $\Gb^{\infty}$. Interestingly, each $\phi_i$ corresponds to $\frac{d \theta_i}{d\vb}$. The full proof is presented in Appendix~\ref{appendix:flow_lemmas}. As already observed from evolution \eqref{eq:flow_evolution}, 
different magnitudes of $\alpha$ can lead to two distinct regimes that are discussed below. We present the main results for each regime. 

\subsection*{$\Vb$-dominated convergence}
For $\alpha<1$ convergence is dominated by $\Vb(t)$ and $\lambda_{\min} (\bLambda(t)) \ge \frac{1}{\alpha^2} \lambda_{\min}(\Vb(t))$. We present the convergence theorem for the $\Vb$-dominated regime here. 
\begin{theorem}[$\Vb$-dominated convergence]
\label{theorem:finite_step_V} 

Suppose a neural network of the form \eqref{eqn:netclass} is initialized as in \eqref{eq:init} with $\alpha \le 1$ and that Assumptions~\ref{as:parallel},\ref{as:mdelta} hold. In addition, suppose the neural network is trained via the regression loss \eqref{eq:regression} with targets $\yb$ satisfying $\|\yb \|_{\infty} = O(1)$. If $m =\Omega\big(n^4 \log(n /\delta) / \lambda_0^4 \big)$,
then with probability $1-\delta$,
\begin{enumerate}
\item For iterations $s = 0,1, \ldots $, the evolution matrix $\bLambda(s)$ satisfies $\lambda_{\min}(\bLambda(s)) \ge \frac{\lambda_0}{2\alpha^2}$. 
\item WN training with gradient descent of step-size $\eta = O\Big(\frac{\alpha^2}{\|\Vb^{\infty}\|_2}\Big)$ converges linearly as
\begin{align*}
    \|\fb(s) - \yb\|_2^2 \le \Big( 1- \frac{\eta \lambda_0}{2\alpha^2}\Big)^s \|\fb(0) - \yb\|_2^2.
\end{align*}
\end{enumerate}
\end{theorem}
The proof of Theorem \ref{theorem:finite_step_V} is presented in Appendix~\ref{sec:finite_step}. 
We will provide a sketch below. We make the following observations about our $\Vb$-dominated convergence result. 

The required over-parametrization $m$ is independent of $\alpha$. Further, the dependence of $m$ on the failure probability is $\log(1/ \delta)$. This improves
previous results that require polynomial dependence of order $\delta^3$. Additionally, we reduce the dependence on the sample size from $n^6$ (as appears in \cite{arora2019fine}) to $n^4 \log(n)$.

In Theorem \ref{theorem:finite_step_V}, smaller $\alpha$ leads to faster convergence, since the convergence is dictated by $\lambda_0/\alpha^2$.  Nonetheless, smaller $\alpha$ is also at the cost of smaller allowed step-sizes, since $\eta = O(\alpha^2 /\|\Vb
^{\infty}\|_2)$. The trade-off between step-size and convergence speed is typical. For example, this is implied in Chizat et al.~\cite{chizat2018note}, where nonetheless the authors point out that for gradient flow training, the increased convergence rate is not balanced by a limitation on the step-size. The works \cite{hoffer2018norm, wu2018wngrad, arora2018theoretical} define an effective step-size (adaptive step-size) $\eta' = \eta/\alpha^2$ to avoid the dependence of $\eta$ on $\alpha$. 

\subsection*{$\Gb$-dominated convergence}
For $\alpha > 1$ our convergence result for the class \eqref{eqn:netclass} is based on monitoring the least eigenvalue of $\Gb(t)$. Unlike $\Vb$-dominated convergence, $\alpha$ does not affect the convergence speed in this regime. 

\begin{theorem}[$\Gb$-dominated convergence]
\label{theorem:finite_step_G}
Suppose a network of the form \eqref{eqn:netclass} is initialized as in \eqref{eq:init} with $\alpha \ge 1$ and that Assumptions~\ref{as:parallel}, \ref{as:mdelta} hold. In addition, suppose the neural network is trained via the regression loss \eqref{eq:regression} with targets $\yb$ satisfying $\|\yb \|_{\infty} = O(1)$. If $m = \Omega \big( \max \big\{n^4 \log(n/ \delta) / \alpha^4 \mu_0^4 , n^2 \log(n /\delta)/\mu_0^2 \big\} \big)$, then with probability $1-\delta$,
\begin{enumerate}
\item For iterations $s = 0,1,\ldots$, the evolution matrix $\bLambda(s)$ satisfies $\lambda_{\min}(\bLambda(s)) \ge \frac{\mu_0}{2}$. 
\item WN training with gradient descent of step-size $\eta = O\Big(\frac{1}{\|\bLambda(t)\|}\Big)$ converges linearly as
\begin{align*}
    \|\fb(s) - \yb\|_2^2 \le \Big( 1- \frac{\eta \mu_0}{2}\Big)^s \|\fb(0) - \yb\|_2^2.
\end{align*}
\end{enumerate}
\end{theorem}
We make the following observations about our $\Gb$-dominated convergence result, and provide a proof sketch further below. 

Theorem \ref{theorem:finite_step_G} holds for $\alpha \ge 1$ so long as 
$m = \Omega \big( \max \big\{n^4 \log(n/ \delta) / \mu_0^4\alpha^4 , n^2 \log(n /\delta)/\mu_0^2 \big\} \big)$. Taking $\alpha =\sqrt{n / \mu_0}$ gives an optimal required over-parametrization of order 
$m = \Omega \big( n^2 \log(n /\delta) /\mu_0^2 \big).$ This significantly improves on previous results \cite{du2018gradient}
for un-normalized training that have dependencies of order $4$ in the least eigenvalue, cubic dependence in $1 /\delta$, and $n^6$ dependence in the number of samples~$n$. In contrast to $\Vb$-dominated convergence, here the rate of convergence $\mu_0$ is independent of $\alpha$ but the over-parametrization $m$ is $\alpha$-dependent. We elaborate on this curious behavior in the next sections.

\paragraph{Proof sketch of main results}
The proof of Theorems \ref{theorem:finite_step_V} and \ref{theorem:finite_step_G} is inspired by a series of works including \cite{du2018gradient,arora2019fine,zhang2019fast, wu2019global, du2018deepgradient}. 
The proof has the following steps: 
($\Ib$)~~We show that at initialization $\Vb(0), \Gb(0)$ can be viewed as empirical estimates of averaged data-dependent kernels $\Vb^{\infty}, \Gb^{\infty}$ that are strictly positive definite under Assumption~\ref{as:parallel}. 
($\Ib\Ib$)~~For each regime, we prove that the corresponding kernel remains positive definite if $\vb_k(t)$ and $g_k(t)$ remain near initialization for each $1 \le k \le m$. 
$(\Ib\Ib\Ib)$~~Given a uniformly positive definite evolution matrix $\bLambda(t)$ and sufficient over-parametrization we show that each neuron, $\vb_k(t), g_k(t)$ remains close to its initialization. 
The full proof is presented in Appendix \ref{appendix:flow_proof} for gradient flow and Appendix \ref{sec:finite_step} for finite-step gradient descent. Next we interpret the main results and discuss how the modified NTK in WN can be viewed as a form of natural gradient.

\paragraph{Connection with natural gradient}
Natural gradient methods define the steepest descent direction in the parameter space of a model from the perspective of function space. 
This amounts to introducing a particular geometry into the parameter space which is reflective of the geometry of the corresponding functions. A re-parametrization of a model, and WN in particular, can also be interpreted as choosing a particular geometry for the parameter space. 
This gives us a perspective from which to study the effects of WN. 
The recent work of \cite{zhang2019fast} studies the effects of natural gradient methods from the lens of the NTK and shows that when optimizing with the natural gradient, one is able to get significantly improved training speed. In particular, using the popular natural gradient method K-FAC improves the convergence speed considerably.

Natural gradients transform the NTK from $\Jb\Jb^\top$ to $\Jb\Gb^\dag \Jb^\top$, where $\Jb$ is the Jacobian with respect to the parameters and $\Gb$ is the metric. The WN re-parametrization transforms the NTK from $\Jb\Jb^\top$ to~$\Jb\Sbb^\top\Sbb \Jb^\top$. To be more precise, denote the un-normalized NTK as $\Hb = \Jb \Jb^{\top}$, where $\Jb$ is the Jacobian matrix for $\xb_1, \dots \xb_n$ written in a compact tensor as $\Jb = \big[\Jb_1, \dots \Jb_n\big]^\top$ with $\Jb_i = \bigg[\pdv{f(\xb_i)}{\wb_1} \dots \pdv{f(\xb_i)}{\wb_m} \bigg]$, where matrix multiplication is a slight abuse of notation. Namely $\Jb \in \RR^{n \times m \times d}$ and we define multiplication of $\Ab\in \RR^{n \times m \times d} \times \Bb \in \RR^{d \times m \times p} \rightarrow \Ab\Bb \in \RR^{n \times p}$ as \begin{align*}
    (\Ab\Bb)_{ij} = \sum_{k=1}^{m} \langle \Ab_{ik:} , \Bb_{: k j} \rangle .
\end{align*}For any re-parametrization $\wb(\rb)$, we have that 
\begin{align*}
    \bLambda = \Kb   \Kb^\top, 
\end{align*}
where $\Kb = \Jb\Sbb^\top$ and $\Sbb$ corresponds to the Jacobian of the re-parametrization $\wb(\rb)$. 
By introducing WN layers the reparameterized NTK is compactly written as 
\begin{align*}
   \bLambda = \Jb \Sbb^\top \Sbb \Jb^\top. 
\end{align*}
Here $\Sbb = [\Sbb_1, \ldots,t \Sbb_m]$ with \begin{align*}
    \Sbb_k = \bigg[ \frac{g_k}{\|\vb_k\|_2}\bigg(\Ib - \frac{\vb_k \vb_k^\top}{\|\vb_k\|_2}\bigg), \frac{\vb_k}{\|\vb_k\|_2} \bigg].
\end{align*}
The term $\Nb(\alpha) := \Sbb \Sbb^\top$ leads to a family of different gradient re-parametrizations depending on $\alpha$. The above representation of the WN NTK is equivalent to $ \bLambda(\alpha) = \frac{1}{\alpha^2}\Vb + \Gb = \Jb \Nb(\alpha) \Jb^\top$. For different initialization magnitudes $\alpha$,~$\Nb(\alpha)$ leads to different NTKs with modified properties. 

For $\alpha = 1$ the term corresponds to training without normalization, yet over $\alpha \in (0, \infty)$,~$\Nb(\alpha)$ leads to a family NTKs with different properties. In addition there exists an $\alpha^*$ that maximizes the convergence rate. Such $\alpha^*$ is either a proper global maximum or is attained at one of $\alpha \rightarrow 0, \alpha \rightarrow \infty$. For the latter, one may fix $\alpha^*$ with $\alpha^* \ll 1$ or $\alpha^* \gg 1$ respectively so that there exists $\alpha^*$ that outpaces un-normalized convergence ($\alpha=1$). This leads to equal or faster convergence of WN as compared with un-normalized training:
\begin{proposition}[Fast Convergence of WN]
\label{theorem:fast_WN} 

Suppose a neural network of the form \eqref{eqn:netclass} is initialized as in \eqref{eq:init} and that Assumptions~\ref{as:parallel},\ref{as:mdelta} hold. In addition, suppose the network is trained via the regression loss \eqref{eq:regression} with targets $\yb$ satisfying $\|\yb \|_{\infty} = O(1)$.
Then, with probability $1-\delta$ over the initialization, there exists $\alpha^*$ such that WN training with $\alpha^*$ initialization leads to faster convergence:
 If $m =\Omega\big(n^4 \log(n /\delta) / \min\{\lambda_0^4, \mu_0^4 \} \big)$,
\begin{enumerate}
\item WN training with gradient descent of step-size $\eta_{\alpha^*} = O\Big(\frac{1}{\|\Vb^{\infty}/(\alpha^*)^2 + \Gb^{\infty} \|_2}\Big)$ converges linearly as
\begin{align*}
&\|\fb(s) - \yb\|_2^2 \le \\
&\bigg( 1- \eta_{\alpha^*} \big(\lambda_0/2(\alpha^*)^2 +\mu_0/2 \big)\bigg)^s \|\fb(0) - \yb\|_2^2.
\end{align*}
\item The convergence rate of WN is faster than un-normalized convergence,
\begin{align*}
    \big( 1-\eta_{\alpha^*}\lambda_{\min}(\bLambda(s))\big) \le \big(1 - \eta \lambda_{\min}(\Hb(s)) \big).
\end{align*}
\end{enumerate}
\end{proposition}
This illustrates the utility of WN from the perspective of the NTK, guaranteeing that there exists an $\alpha^*$ that leads to faster convergence in \emph{finite-step} gradient descent as compared with un-normalized training. 

\section{Related Work}\label{sec:related_works}
\paragraph{Normalization methods theory}
A number of recent works attempt to explain the dynamics and utility of various normalization methods in deep learning. 
The original works on BN \cite{ioffe2015batch} and WN \cite{salimans2016weight} suggest that normalization procedures improve training by fixing the intermediate layers' output distributions. 
The works of \citet{bjorck2018understanding} and \citet{santurkar2018does} argue that BN may improve optimization by improving smoothness of the Hessian of the loss, therefore allowing for larger step-sizes with reduced instability. \citet{hoffer2017train} showed that the effective step-size in BN is divided by the magnitude of the weights. This followed the work on WNgrad \cite{wu2018wngrad} that introduces an adaptive step-size algorithm based on this fact. Following the intuition of WNGrad, \citet{arora2018theoretical} proved that for smooth loss and network functions, the diminishing ``effective step-size'' of normalization methods leads to convergence with optimal convergence rate for properly initialized step-sizes. The work of \citet{kohler2018towards} explains the accelerated convergence of BN from a ``length-direction decoupling'' perspective. The authors along with \citet{cai2019quantitative} analyze the linear least squares regime, with \citet{kohler2018towards} presenting a bisection method for finding the optimal weights. Robustness and regularization of Batch Normalization is investigated by \citet{luo2018understanding} and improved generalization is analyzed empirically. 
Shortly after the original work of WN, \cite{yoshida2017statistical} showed that for a single precptron WN may speed-up training and emphasized the importance of the norm of the initial weights. 
Additional stability properties were studied by \citet{yang2019mean} via mean-field analysis. The authors show that gradient instability is inevitable even with BN as the number of layers increases; this is in agreement with \citet{balduzzi2017shattered} for networks with residual connections. 
The work of \citet{NIPS2019_9272} suggests initialization strategies for WN and derives lower bounds on the width to guarantee same order gradients across the layers.

\paragraph{Over-parametrized neural networks}
There has been a significant amount of recent literature studying the convergence of un-normalized over-parametrized neural networks. In the majority of these works the analysis relies on the width of the layers. 
These include 2-layer networks trained with Gaussian inputs and outputs from a teacher network \cite{tian2017analytical,li2017convergence} and \cite{du2017convolutional} (with WN). 
Assumptions on the data distribution are relaxed in \cite{du2018gradient} and the works that followed \cite{zhang2019fast,arora2019fine,wu2019global}. Our work is inspired by the mechanism presented in this chain of works.  \citet{wu2019global} extend convergence results to adaptive step-size methods and propose AdaLoss. Recently, the global convergence of over-parameterized neural networks was also extended to deep architectures \cite{du2018deepgradient,allen2019convergence,zou2018stochastic, zou2019improved}. In the context of the NTK, \citet{zhang2019fast} have proved fast convergence of neural networks trained with natural gradient methods and the K-FAC approximation \cite{martens2015optimizing}. In the over-parameterized regimes, \citet{arora2019fine} develop generalization properties for the networks of the form \eqref{eq:un-norm-net-class}. In addition, in the context of generalization, \citet{allen2018learning} illustrates good generalization for deep neural networks trained with gradient descent. \citet{cao2019generalization} and \cite{cao2019generalizationsgd} derive generalization error bounds of gradient descent and stochastic gradient descent for learning over-parametrization deep ReLU neural networks.

\section{Discussion} \label{section:discussion}
Dynamic normalization is the most common optimization set-up of current deep learning models, yet understanding the convergence of such optimization methods is still an open problem. In this work we present a proof giving sufficient conditions for convergence of dynamically normalized 2-layer ReLU networks trained with gradient descent. To the best of our knowledge this is the first proof showcasing convergence of gradient descent training of neural networks with dynamic normalization and general data, where the objective function is non-smooth and non-convex. To understand the canonical behavior of each normalization layer, we study the shallow neural network case, that enables us to focus on a single layer and illustrate the dynamics of weight normalization. Nonetheless, we believe that using the techniques presented in \cite{allen2019convergence,du2018deepgradient} can extend the proofs to the deep network settings. Through our analysis notion of ``length-direction decoupling'' is clarified by the neural tangent kernel $\bLambda(t)$ that naturally separates in our analysis into
``length'', $\Gb(t)$, and ``direction'', $\Vb(t) / \alpha^2$, components. For $\alpha=1$ the decomposition initially matches un-normalized training.
Yet we discover that in general, normalized training with gradient descent leads to 2 regimes dominated by different pieces of the neural tangent kernel. Our improved analysis is able to reduce the amount of over-parametrization that was needed in previous convergence works in the un-normalized setting and in the $\Gb$-dominated regime, we prove convergence with a significantly lower amount of over-parametrization as compared with un-normalized training. 
\raggedbottom

\noindent\textbf{Acknowledgement} 
YD has been supported by the National Science Foundation under Graduate Research Fellowship Grant No. DGE-1650604. QG was supported in part by the National Science Foundation CAREER Award IIS-1906169, BIGDATA IIS-1855099, and Salesforce Deep Learning Research Award. This project has received funding from the European Research Council (ERC) under the European Union's Horizon 2020 research and innovation programme (grant agreement n\textsuperscript{o} 757983). 

\bibliography{paper}

\begin{thebibliography}{38}
\providecommand{\natexlab}[1]{#1}
\providecommand{\url}[1]{\texttt{#1}}
\expandafter\ifx\csname urlstyle\endcsname\relax
  \providecommand{\doi}[1]{doi: #1}\else
  \providecommand{\doi}{doi: \begingroup \urlstyle{rm}\Url}\fi

\bibitem[Allen-Zhu et~al.(2019{\natexlab{a}})Allen-Zhu, Li, and
  Liang]{allen2018learning}
Zeyuan Allen-Zhu, Yuanzhi Li, and Yingyu Liang.
\newblock Learning and generalization in overparameterized neural networks,
  going beyond two layers.
\newblock In \emph{Advances in Neural Information Processing Systems 32}, pages
  6158--6169. 2019{\natexlab{a}}.

\bibitem[Allen-Zhu et~al.(2019{\natexlab{b}})Allen-Zhu, Li, and
  Song]{allen2019convergence}
Zeyuan Allen-Zhu, Yuanzhi Li, and Zhao Song.
\newblock A convergence theory for deep learning via over-parameterization.
\newblock In \emph{Proceedings of the 36th International Conference on Machine
  Learning}, volume~97 of \emph{Proceedings of Machine Learning Research},
  pages 242--252. PMLR, 2019{\natexlab{b}}.

\bibitem[Arora et~al.(2019{\natexlab{a}})Arora, Du, Hu, Li, and
  Wang]{arora2019fine}
Sanjeev Arora, Simon Du, Wei Hu, Zhiyuan Li, and Ruosong Wang.
\newblock Fine-grained analysis of optimization and generalization for
  overparameterized two-layer neural networks.
\newblock In \emph{Proceedings of the 36th International Conference on Machine
  Learning}, volume~97 of \emph{Proceedings of Machine Learning Research},
  pages 322--332. PMLR, 2019{\natexlab{a}}.

\bibitem[Arora et~al.(2019{\natexlab{b}})Arora, Li, and
  Lyu]{arora2018theoretical}
Sanjeev Arora, Zhiyuan Li, and Kaifeng Lyu.
\newblock Theoretical analysis of auto rate-tuning by batch normalization.
\newblock In \emph{International Conference on Learning Representations},
  2019{\natexlab{b}}.
\newblock URL \url{https://openreview.net/forum?id=rkxQ-nA9FX}.

\bibitem[Arpit et~al.(2019)Arpit, Campos, and Bengio]{NIPS2019_9272}
Devansh Arpit, V\'{\i}ctor Campos, and Yoshua Bengio.
\newblock How to initialize your network? robust initialization for weightnorm
  \&amp; resnets.
\newblock In H.~Wallach, H.~Larochelle, A.~Beygelzimer, F.~d~Alch\'{e}-Buc,
  E.~Fox, and R.~Garnett, editors, \emph{Advances in Neural Information
  Processing Systems 32}, pages 10902--10911. Curran Associates, Inc., 2019.

\bibitem[Ba et~al.(2016)Ba, Kiros, and Hinton]{ba2016layer}
Jimmy~Lei Ba, Jamie~Ryan Kiros, and Geoffrey~E Hinton.
\newblock Layer normalization.
\newblock \emph{Deep Learning Symposium, NIPS-2016}, 2016.

\bibitem[Balduzzi et~al.(2017)Balduzzi, Frean, Leary, Lewis, Ma, and
  McWilliams]{balduzzi2017shattered}
David Balduzzi, Marcus Frean, Lennox Leary, JP~Lewis, Kurt Wan-Duo Ma, and
  Brian McWilliams.
\newblock The shattered gradients problem: If resnets are the answer, then what
  is the question?
\newblock In \emph{Proceedings of the 34th International Conference on Machine
  Learning}, volume~70 of \emph{Proceedings of Machine Learning Research},
  pages 342--350. JMLR. org, 2017.

\bibitem[Bjorck et~al.(2018)Bjorck, Gomes, Selman, and
  Weinberger]{bjorck2018understanding}
Nils Bjorck, Carla~P Gomes, Bart Selman, and Kilian~Q Weinberger.
\newblock Understanding batch normalization.
\newblock In \emph{Advances in Neural Information Processing Systems 31}, pages
  7694--7705. 2018.

\bibitem[Cai et~al.(2019)Cai, Li, and Shen]{cai2019quantitative}
Yongqiang Cai, Qianxiao Li, and Zuowei Shen.
\newblock A quantitative analysis of the effect of batch normalization on
  gradient descent.
\newblock In \emph{International Conference on Machine Learning}, pages
  882--890, 2019.

\bibitem[Cao and Gu(2019)]{cao2019generalizationsgd}
Yuan Cao and Quanquan Gu.
\newblock Generalization bounds of stochastic gradient descent for wide and
  deep neural networks.
\newblock In \emph{Advances in Neural Information Processing Systems 32}, pages
  10836--10846. 2019.

\bibitem[Cao and Gu(2020)]{cao2019generalization}
Yuan Cao and Quanquan Gu.
\newblock Generalization error bounds of gradient descent for learning
  over-parameterized deep {ReLU} networks.
\newblock In \emph{AAAI}, 2020.

\bibitem[Chizat et~al.(2019)Chizat, Oyallon, and Bach]{chizat2018note}
L\'{e}na\"{\i}c Chizat, Edouard Oyallon, and Francis Bach.
\newblock On lazy training in differentiable programming.
\newblock In \emph{Advances in Neural Information Processing Systems 32}, pages
  2937--2947. 2019.

\bibitem[Du et~al.(2019{\natexlab{a}})Du, Lee, Li, Wang, and
  Zhai]{du2018deepgradient}
Simon Du, Jason Lee, Haochuan Li, Liwei Wang, and Xiyu Zhai.
\newblock Gradient descent finds global minima of deep neural networks.
\newblock In \emph{Proceedings of the 36th International Conference on Machine
  Learning}, volume~97 of \emph{Proceedings of Machine Learning Research},
  pages 1675--1685, Long Beach, California, USA, 09--15 Jun 2019{\natexlab{a}}.
  PMLR.

\bibitem[Du et~al.(2018)Du, Lee, and Tian]{du2017convolutional}
Simon~S. Du, Jason~D. Lee, and Yuandong Tian.
\newblock When is a convolutional filter easy to learn?
\newblock In \emph{International Conference on Learning Representations}, 2018.
\newblock URL \url{https://openreview.net/forum?id=SkA-IE06W}.

\bibitem[Du et~al.(2019{\natexlab{b}})Du, Zhai, Poczos, and
  Singh]{du2018gradient}
Simon~S. Du, Xiyu Zhai, Barnabas Poczos, and Aarti Singh.
\newblock Gradient descent provably optimizes over-parameterized neural
  networks.
\newblock In \emph{International Conference on Learning Representations},
  2019{\natexlab{b}}.
\newblock URL \url{https://openreview.net/forum?id=S1eK3i09YQ}.

\bibitem[Gitman and Ginsburg(2017)]{gitman2017comparison}
Igor Gitman and Boris Ginsburg.
\newblock Comparison of batch normalization and weight normalization algorithms
  for the large-scale image classification.
\newblock \emph{arXiv preprint arXiv:1709.08145}, 2017.

\bibitem[Hoffer et~al.(2017)Hoffer, Hubara, and Soudry]{hoffer2017train}
Elad Hoffer, Itay Hubara, and Daniel Soudry.
\newblock Train longer, generalize better: closing the generalization gap in
  large batch training of neural networks.
\newblock In \emph{Advances in Neural Information Processing Systems 30}, pages
  1731--1741. 2017.

\bibitem[Hoffer et~al.(2018)Hoffer, Banner, Golan, and Soudry]{hoffer2018norm}
Elad Hoffer, Ron Banner, Itay Golan, and Daniel Soudry.
\newblock Norm matters: efficient and accurate normalization schemes in deep
  networks.
\newblock In \emph{Advances in Neural Information Processing Systems 31}, pages
  2160--2170. 2018.

\bibitem[Ioffe and Szegedy(2015)]{ioffe2015batch}
Sergey Ioffe and Christian Szegedy.
\newblock Batch normalization: Accelerating deep network training by reducing
  internal covariate shift.
\newblock In \emph{Proceedings of the 32nd International Conference on Machine
  Learning}, volume~37 of \emph{Proceedings of Machine Learning Research},
  pages 448--456. PMLR, 2015.

\bibitem[Jacot et~al.(2018)Jacot, Gabriel, and Hongler]{jacot2018neural}
Arthur Jacot, Franck Gabriel, and Clement Hongler.
\newblock Neural tangent kernel: Convergence and generalization in neural
  networks.
\newblock In \emph{Advances in Neural Information Processing Systems 31}, pages
  8571--8580. 2018.

\bibitem[Kohler et~al.(2019)Kohler, Daneshmand, Lucchi, Hofmann, Zhou, and
  Neymeyr]{kohler2018towards}
Jonas Kohler, Hadi Daneshmand, Aurelien Lucchi, Thomas Hofmann, Ming Zhou, and
  Klaus Neymeyr.
\newblock Exponential convergence rates for batch normalization: The power of
  length-direction decoupling in non-convex optimization.
\newblock In \emph{Proceedings of Machine Learning Research}, volume~89 of
  \emph{Proceedings of Machine Learning Research}, pages 806--815. PMLR, 2019.

\bibitem[Li and Liang(2018)]{li2018learning}
Yuanzhi Li and Yingyu Liang.
\newblock Learning overparameterized neural networks via stochastic gradient
  descent on structured data.
\newblock In \emph{Advances in Neural Information Processing Systems 31}, pages
  8157--8166. 2018.

\bibitem[Li and Yuan(2017)]{li2017convergence}
Yuanzhi Li and Yang Yuan.
\newblock Convergence analysis of two-layer neural networks with {ReLU}
  activation.
\newblock In \emph{Advances in Neural Information Processing Systems 30}, pages
  597--607. 2017.

\bibitem[Luo et~al.(2018)Luo, Wang, Shao, and Peng]{luo2018understanding}
Ping Luo, Xinjiang Wang, Wenqi Shao, and Zhanglin Peng.
\newblock Understanding regularization in batch normalization.
\newblock \emph{arXiv preprint arXiv:1809.00846}, 2018.

\bibitem[Martens and Grosse(2015)]{martens2015optimizing}
James Martens and Roger Grosse.
\newblock Optimizing neural networks with {K}ronecker-factored approximate
  curvature.
\newblock In \emph{Proceedings of the 32nd International Conference on Machine
  Learning}, volume~37 of \emph{Proceedings of Machine Learning Research},
  pages 2408--2417. PMLR, 2015.

\bibitem[Nair and Hinton(2010)]{nair2010rectified}
Vinod Nair and Geoffrey~E Hinton.
\newblock Rectified linear units improve restricted {B}oltzmann machines.
\newblock In \emph{Proceedings of the 27th international conference on machine
  learning (ICML-10)}, pages 807--814, 2010.

\bibitem[Oymak and Soltanolkotabi(2019)]{oymak2019towards}
Samet Oymak and Mahdi Soltanolkotabi.
\newblock Towards moderate overparameterization: global convergence guarantees
  for training shallow neural networks.
\newblock \emph{arXiv preprint arXiv:1902.04674}, 2019.

\bibitem[Salimans and Kingma(2016)]{salimans2016weight}
Tim Salimans and Durk~P Kingma.
\newblock Weight normalization: A simple reparameterization to accelerate
  training of deep neural networks.
\newblock In \emph{Advances in Neural Information Processing Systems 29}, pages
  901--909. 2016.

\bibitem[Santurkar et~al.(2018)Santurkar, Tsipras, Ilyas, and
  Madry]{santurkar2018does}
Shibani Santurkar, Dimitris Tsipras, Andrew Ilyas, and Aleksander Madry.
\newblock How does batch normalization help optimization?
\newblock In \emph{Advances in Neural Information Processing Systems 31}, pages
  2483--2493. 2018.

\bibitem[Tian(2017)]{tian2017analytical}
Yuandong Tian.
\newblock An analytical formula of population gradient for two-layered {ReLU}
  network and its applications in convergence and critical point analysis.
\newblock In \emph{Proceedings of the 34th International Conference on Machine
  Learning-Volume 70}, pages 3404--3413. JMLR. org, 2017.

\bibitem[Vershynin(2018)]{vershynin2018high}
Roman Vershynin.
\newblock \emph{High-dimensional probability: An introduction with applications
  in data science}, volume~47.
\newblock Cambridge University Press, 2018.

\bibitem[Wu et~al.(2018)Wu, Ward, and Bottou]{wu2018wngrad}
Xiaoxia Wu, Rachel Ward, and L{\'e}on Bottou.
\newblock {WNGrad}: Learn the learning rate in gradient descent.
\newblock \emph{arXiv preprint arXiv:1803.02865}, 2018.

\bibitem[Wu et~al.(2019)Wu, Du, and Ward]{wu2019global}
Xiaoxia Wu, Simon~S Du, and Rachel Ward.
\newblock Global convergence of adaptive gradient methods for an
  over-parameterized neural network.
\newblock \emph{arXiv preprint arXiv:1902.07111}, 2019.

\bibitem[Yang et~al.(2019)Yang, Pennington, Rao, Sohl-Dickstein, and
  Schoenholz]{yang2019mean}
Greg Yang, Jeffrey Pennington, Vinay Rao, Jascha Sohl-Dickstein, and Samuel~S.
  Schoenholz.
\newblock A mean field theory of batch normalization.
\newblock In \emph{International Conference on Learning Representations}, 2019.
\newblock URL \url{https://openreview.net/forum?id=SyMDXnCcF7}.

\bibitem[Yoshida et~al.(2017)Yoshida, Karakida, Okada, and
  Amari]{yoshida2017statistical}
Yuki Yoshida, Ryo Karakida, Masato Okada, and Shun-ichi Amari.
\newblock Statistical mechanical analysis of online learning with weight
  normalization in single layer perceptron.
\newblock \emph{Journal of the Physical Society of Japan}, 86\penalty0
  (4):\penalty0 044002, 2017.

\bibitem[Zhang et~al.(2019)Zhang, Martens, and Grosse]{zhang2019fast}
Guodong Zhang, James Martens, and Roger~B Grosse.
\newblock Fast convergence of natural gradient descent for over-parameterized
  neural networks.
\newblock In \emph{Advances in Neural Information Processing Systems 32}, pages
  8082--8093. 2019.

\bibitem[Zou and Gu(2019)]{zou2019improved}
Difan Zou and Quanquan Gu.
\newblock An improved analysis of training over-parameterized deep neural
  networks.
\newblock In \emph{Advances in Neural Information Processing Systems 32}, pages
  2055--2064. 2019.

\bibitem[Zou et~al.(2020)Zou, Cao, Zhou, and Gu]{zou2018stochastic}
Difan Zou, Yuan Cao, Dongruo Zhou, and Quanquan Gu.
\newblock Gradient descent optimizes over-parameterized deep {ReLU} networks.
\newblock \emph{Machine Learning}, 109\penalty0 (3):\penalty0 467--492, 2020.
\newblock \doi{10.1007/s10994-019-05839-6}.
\newblock URL \url{https://doi.org/10.1007/s10994-019-05839-6}.

\end{thebibliography}
\bibliographystyle{plainnat}
\pagebreak
\onecolumn
\appendix

\section*{Appendix}
We present the detailed proofs of the main results of the paper below. The appendix is organized as follows. We provide proofs to the simple propositions regarding the NTK presented in the paper in Appendix \ref{sec:WN_proofs}, and prove the main results for $\Vb$-dominated and $\Gb$-dominated convergence in the settings of gradient flow and gradient descent in Appendices \ref{appendix:flow_proof} and \ref{sec:finite_step}. The proofs for gradient flow and gradient descent share the same main idea, yet the proof for gradient descent has a considerate number of additional technicalities. In Appendices \ref{appendix:flow_lemmas} and \ref{appendix:finite_step_lemmas} we prove the lemmas used in the analysis of Appendices \ref{appendix:flow_proof} and \ref{sec:finite_step} respectively.

\section{Weight Normalization Dynamics Proofs}\label{sec:WN_proofs}
In this section we provide proofs for Proposition~\ref{rem:VGH}, 
which describes the relation between vanilla and WeightNorm NTKs and Observation \ref{obs:inv} of the paper. \par
\begin{proofProp}{\ref{rem:VGH}}
We would like to show that $\Vb(0) +\Gb(0) = \Hb(0)$. For each entry, consider 
\begin{align*}
(\Vb(0) + \Gb(0))_{ij} = \frac{1}{m} \sum_{k=1}^{m} \big \langle \xb_i^{\vb_k(0)^\perp},~ {\xb_j}^{\vb_k(0)^\perp} \big \rangle \ind_{ik}(0)\ind_{jk}(0) +   \frac{1}{m} \sum_{k=1}^{m}\big \langle \xb_i^{\vb_k(0)},~{\xb_j}^{\vb_k(0)} \big \rangle \ind_{ik}(0)\ind_{jk}(0) . 
\end{align*}
Note that 
\begin{align*}\big \langle \xb_i,~\xb_j\big \rangle = \big \langle \xb_i^{\vb_k(0)} + \xb_i^{\vb_k(0)^\perp} ,~\xb_j^{\vb_k(0)} + \xb_j^{\vb_k(0)^\perp} \big \rangle =  \big \langle \xb_i^{\vb_k(0)^\perp},~{\xb_j}^{\vb_k(0)^\perp} \big \rangle + \big \langle \xb_i^{\vb_k(0)},~{\xb_j}^{\vb_k(0)} \big \rangle . 
\end{align*}
This gives 
\begin{align*}
(\Vb(0) + \Gb(0))_{ij}  = \frac{1}{m}\sum_{k=1}^{m} \big \langle \xb_i,~ \xb_j \big \rangle\ind_{ik}(0) \ind_{jk}(0) = \Hb_{ij}(0) 
\end{align*}
which proves the claim. 
\end{proofProp}

\begin{proofObs}{\ref{obs:inv}}\label{proof:obs_inv}%
We show that the initialization of the network is independent of $\alpha$. Take $\alpha, \beta > 0$, and for each $k$, initialize $\vb_k^\alpha, \vb_k^\beta$ as
\begin{align*}
    \vb^{\alpha}_k(0) \sim N(0, \alpha^2 \Ib),\quad \vb^{\beta}_k(0) \sim N(0, \beta^2 \Ib). 
\end{align*} 
Then 
\begin{align*}
\frac{\vb^{\alpha}_k(0)}{\|\vb^{\alpha}_k(0)\|_2} \sim \frac{\vb^{\beta}_k(0)}{\|\vb^{\beta}_k(0)\|_2} \sim ~\text{Unif}(\mathcal{S}^{d-1})~~~\text{(in distribution)}. 
\end{align*}
Hence the distribution of each neuron $\sigma \big( \frac{\vb_k(0)}{\|\vb_k(0)\|_2} \big)$ at initialization is independent of $\alpha$. Next for $g_k(0)$, we note that
\begin{align*}
\|\vb_k^{\alpha}(0) \|_2\sim  \frac{\alpha}{\beta}\|\vb_k^{\beta}(0)\|_2.
\end{align*}
Initializing $g^\alpha_k(0), g^\beta_k(0)$  as in \eqref{eq:init},
\begin{align*} 
g_k^\alpha(0) = \frac{\|\vb_k(0)\|_2}{\alpha},\quad g_k^\beta(0) = \frac{\|\vb_k(0)\|_2}{\beta},
\end{align*}
gives
\begin{align*}
g_k^\alpha(0), \quad g_k^\beta(0) \sim \chi_d,\quad \text{and }~\frac{g^\alpha_k(0) \vb^\alpha_k(0)}{\|\vb^\alpha_k(0)\|_2} \sim \frac{g_k^\beta(0) \vb_k^\beta(0)}{\|\vb_k^\beta(0)\|_2} \sim N(0, \Ib), 
\end{align*}
for all $\alpha, \beta$. 
This shows that the network initialization is independent of $\alpha$ and is equivalent to the initialization of the un-normalized setting. Similarly, inspecting the terms in the summands of $\Vb(0), \Gb(0)$ shows that they are also independent of $\alpha$. 
For 
\begin{align*}
    \Vb_{ij}(0) =  \frac{1}{m} \sum_{k=1}^{m}\ind_{ik}(0) \ind_{jk}(0)\bigg(\frac{\alpha c_k\cdot g_k(0)}{\|\vb_k(0)\|_2}\bigg)^{2} \big \langle \xb_i ^{\vb_k(0)^\perp},~ \xb_j^{\vb_k(0)^\perp} \big \rangle
    \end{align*}
the terms $\ind_{ik}(0)$, $\xb_i^{\vb_k(0)^\perp}$ are independent of scale, and the fraction in the summand is identically $1$. $\Gb(0)$ defined as
\begin{align*}
       \Gb_{ij}(0) =  \frac{1}{m} \sum_{k=1}^{m}\ind_{ik}(0) \ind_{jk}(0) \big \langle \xb_i ^{\vb_k(0)},~ \xb_j^{\vb_k(0)} \big \rangle
\end{align*}
is also invariant of scale since the projection onto a vector direction $\vb_k(0)$ is independent of scale.
\end{proofObs}

Before we move forward we highlight some of the challenges of the WN proof.
\paragraph{Distinctive aspects of the WN analysis}
The main idea of our proof are familiar and structured similarly to the work by \citet{du2018gradient} on the un-normalized setting. However, the majority of the proofs are modified significantly to account for WN. To the best of our knowledge, the finite-step analysis that we present in Appendix~\ref{sec:finite_step} is entirely new, incorporating updates of both $\vb$ and $g$. The proof of Theorem~\ref{theorem:finite_step_proof} is crucially dependent on the geometry of WN gradient descent and the orthogonality property, in particular \eqref{eq:inc}. 
Updates of the weights in both the numerator and denominator require additional analysis that is presented in Lemma~\ref{lemma:mdelta}. 
In Appendix~\ref{appendix:finite_step_lemmas} we prove Theorems~\ref{theorem:finite_step_V}, \ref{theorem:finite_step_G} based on the general Theorem~\ref{theorem:finite_step_proof} and Property~\ref{property:cond} which is based on new detailed decomposition of the finite-step difference between iterations. 
In contrast to the un-normalized setting, the auxiliary matrices $\Vb^{\infty}, \Gb^{\infty}$ that we have in the WN analysis are not piece-wise constant in $\vb$. 
To prove they are positive definite, we prove Lemma~\ref{lemma:hau} based on two new constructive arguments. 
We develop the technical Lemma~\ref{lemma:bern} and utilize Bernstein's inequality to reduce the amount of required over-parametrization in our final bounds on the width $m$. 
The amount of over-parameterization in relation to the sample size $n$ is reduced (from $n^6$ to $n^4$) through more careful arguments in Lemmas \ref{lemma:rec} and \ref{lemma:ht}, which introduce an intermediate matrix $\hat{\Vb}(t)$ and follow additional geometrical identities. 
Lemma~\ref{lemma:init} reduces the polynomial dependence on the failure probability $\delta$ to logarithmic dependence based on sub-Gaussian concentration. 
The denominator in the WN architecture necessities worst bound analysis which we handle in Lemma~\ref{lemma:mdelta} that is used extensively throughout the proofs. 
%Additionally we present the proof of Proposition \ref{theorem:fast_WN} of faster convergence in \ref{appendix:finite_step_lemmas}.

\section{Convergence Proof for Gradient Flow }\label{appendix:flow_proof}
In this section we derive the convergence results for gradient flow.

The main results are analogous to Theorems \ref{theorem:finite_step_V}, \ref{theorem:finite_step_G} but by considering gradient flow instead of gradient descent the proofs are simplified. In Appendix \ref{sec:finite_step} we prove the main results from Section \ref{section:main} (Theorem \ref{theorem:finite_step_V}, \ref{theorem:finite_step_G}) for finite step gradient descent.

We state our convergence results for gradient flow.
\begin{theorem}[$\Vb$-dominated convergence]\label{theorem:flow_proof_V}
Suppose a network from the class \eqref{eqn:netclass} is initialized as in \eqref{eq:init} with $\alpha < 1$ and that assumptions \ref{as:parallel},\ref{as:mdelta} hold. In addition, suppose the neural network is trained via the regression loss \eqref{eq:regression} with target $\yb$ satisfying $\|\yb \|_{\infty} = O(1)$. Then if $m =\Omega\big(n^4 \log(n /\delta) / \lambda_0^4  \big)$,
WeightNorm training with gradient flow converges at a linear rate, with probability $1-\delta$, as
\begin{align*}
\|\fb (t) - \yb\|^{2}_{2} \le \exp( -\lambda_0 t/{\alpha^2} )\|\fb (0) - \yb\|^2_2.
\end{align*}
\end{theorem}
This theorem is analogous to Theorem \ref{theorem:finite_step_V} but since here, the settings are of gradient flow there is no mention of the step-size. It is worth noting that smaller $\alpha$ leads to faster convergence and appears to not affect the other hypotheses of the flow theorem. This ``un-interuptted'' fast convergence behavior does not extend to finite-step gradient descent where the increased convergence rate is balanced by decreasing the allowed step-size.

The second main result for gradient flow is for $\Gb$-dominated convergence.
\begin{theorem}[$\Gb$-dominated convergence]\label{theorem:flow_proof_G}
Suppose a network from the class \eqref{eqn:netclass} is initialized as in \eqref{eq:init} with $\alpha > 1$ and that assumptions \ref{as:parallel}, \ref{as:mdelta} hold. In addition, suppose the neural network is trained on the regression loss \eqref{eq:regression} with target $\yb$ satisfying $\|\yb \|_{\infty} = O(1)$. Then if $m = \Omega \big( \max \big\{n^4 \log(n/ \delta) / \alpha^4 \mu_0^4 , n^2 \log(n /\delta)/\mu_0^2 \big\} \big)$,
WeightNorm training with gradient flow converges at a linear rate, with probability $1-\delta$, as
\begin{align*}
\|\fb (t) - \yb\|^{2}_{2} \le \exp(-\mu_0 t )\|\fb (0) - \yb\|^2_2.
\end{align*}
\end{theorem}

\subsection{Proof Sketch}
To prove the results above we follow the steps introduced in the proof sketch of Section \ref{section:main}. The main idea of the proofs for $\Vb$ and $\Gb$ dominated convergence are analogous and a lot of the proofs are based of \citet{du2018gradient}. We show that in each regime, we attain linear convergence by proving that the least eigenvalue of the evolution matrix $\bLambda(t)$ is strictly positive. For the $\Vb$-dominated regime we lower bound the least eigenvalue of $\bLambda(t)$ as $\lambda_{\min}(\bLambda(t)) \ge \lambda_{\min}(\Vb(t))/\alpha^2$ and in the $\Gb$-dominated regime we lower bound the least eigenvalue as $\lambda_{\min}(\bLambda(t)) \ge \lambda_{\min}(\Gb(t))$.

The main part of the proof is showing that $\lambda_{\min}(\Vb(t)), \lambda_{\min}(\Gb(t))$ stay uniformly positive. We use several lemmas to show this claim.

In each regime, we first show that at initialization the kernel under consideration, $\Vb(0)$ or $\Gb(0)$, has a positive least eigenvalue. This is shown via concentration to an an auxiliary kernel (Lemmas \ref{lemma:hzero}, \ref{lemma:Gzero}), and showing that the auxiliary kernel is also strictly positive definite (Lemma \ref{lemma:hau}).

\begin{lemma} \label{lemma:hzero}Let $\Vb(0)$ and $ \Vb^\infty$ be defined as in \eqref{eq:defineV} and \eqref{eq:DefineVinf}, assume  the network width $m$ satisfies $m= \Omega \big( \frac{n^2 \log(n/\delta) }{\lambda_0^2}\big)$. Then with probability $1-\delta$,
\begin{align*}\|\Vb(0) - \Vb^{\infty} \|_2 \le \frac{\lambda_0}{4}.
\end{align*}
\end{lemma}

\begin{lemma} \label{lemma:Gzero}
Let $\Gb(0)$ and $ \Gb^\infty$ be defined as in \eqref{eq:defineG} and \eqref{eq:DefineGinf}, assume $m$ satisfies $m = \Omega \big( \frac{n^2\log(n /\delta)}{\mu_0 ^2} \big)$. Then with probability $1-\delta$,
\begin{align*}\|\Gb(0) - \Gb^{\infty} \|_2 \le \frac{\mu_0}{4}.
\end{align*}
\end{lemma}

After showing that $\Vb(0), \Gb(0)$ have a positive least-eigenvalue we show that $\Vb(t), \Gb(t)$ maintain this positive least eigenvalue during training. This part of the proof depends on the over-parametrization of the networks. The main idea is showing that if the individual parameters $\vb_k(t), g_k(t)$ do not change too much during training, then $\Vb(t), \Gb(t)$ remain close enough to $\Vb(0), \Gb(0)$ so that they are still uniformly strictly positive definite. We prove the results for $\Vb(t)$ and $\Gb(t)$ separately since each regime imposes different restrictions on the trajectory of the parameters.

For now, in Lemmas \ref{lemma:rec}, \ref{lemma:Vt}, \ref{lemma:Gt}, we make assumptions on the parameters of the network not changing ``too much''; later we show that this holds and is the result of over-parametrization. Specifically, over-parametrization ensures that the parameters stay at a small maximum distance from their initialization. 

\paragraph{$\Vb$-dominated convergence}
To prove the least eigenvalue condition on $\Vb(t)$, we introduce the surrogate Gram matrix $\hat{\Vb}(t)$ defined entry-wise as
\begin{align}\label{eq:define_Rv_Rg}
\hat{\Vb}_{ij}(t) = \frac{1}{m} \sum_{k = 1}^{m} \big \langle \xb_i^{\vb_k(t)^{\perp}},~ \xb_j^{\vb_k(t)^{\perp}} \big \rangle \ind_{ik}(t) \ind_{jk}(t).
\end{align}

This definition aligns with $\Vb(t)$ if we replace the scaling term $\big(\frac{\alpha c_k g_k(t)}{\|\vb_k(t)\|_2} \big)^2$ in each term in the sum $\Vb_{ij}(t)$ by $1$. 

To monitor $\Vb(t) -\Vb(0)$ we consider $\hat{\Vb}(t) - \Vb(0)$ and $\Vb(t) - \hat{\Vb}(t)$ in Lemmas \ref{lemma:rec} and \ref{lemma:Vt} respectively:

\begin{lemma}[Rectifier sign-changes]\label{lemma:rec}
Suppose $\vb_1(0), \ldots, \vb_k(0)$ are sampled i.i.d.\ as \eqref{eq:init}. In addition assume we have $m = \Omega \big( \frac{(m /\delta)^{1 /d}n \log(n / \delta) }{\alpha \lambda_0}\big)$ and  $\| \vb_k(t) - \vb_k(0)\|_2 \le  \frac{ \alpha \lambda_0}{96n(m /\delta)^{1 / d}} \eqqcolon R_v $. Then with probability $1-\delta$, 
\begin{align*}
\|\hat{\Vb}(t) - \Vb(0)\|_2 \le \frac{\lambda_0}{8}.\end{align*}
\end{lemma}

\begin{lemma} \label{lemma:ht}
Define 
\begin{align}\label{eq:neededR}
R_g = \frac{\lambda_0}{48n(m/ \delta)^{1 /d}}, \quad R_v = \frac{\alpha \lambda_0  }{96n
(m/ \delta)^{1 /d}}. 
\end{align}

Suppose the conditions of Lemma \ref{lemma:rec} hold, and that $\|\vb_k(t) -\vb_k(0) \|_2 \le R_v$,  $\|g_k(t) -g_k(0) \|_2 \le R_g$ for all $1 \le k \le m$. Then with probability $1-\delta$,
 \begin{align*}\|\Vb(t) - \Vb(0)\|_2 \le \frac{\lambda_0}{4}. \end{align*}
 \label{lemma:Vt}
\end{lemma}

\paragraph{$\Gb$-dominated convergence}
We ensure that $\Gb(t)$ stays uniformly positive definite if the following hold.
\begin{lemma}\label{lemma:Gt}
Given $\vb_1(0), \ldots, \vb_k(0)$ generated i.i.d.\ as in \eqref{eq:init}, suppose that for each $k$, $\| \vb_k(t) - \vb_k(0) \|_2 \le \frac{\sqrt{2\pi} \alpha \mu_0 }{8 n (m /\delta)^{1/d}} \eqqcolon \tilde{R_v}$,
then with probability $1-\delta$,
\begin{align*}
    \|\Gb(t) - \Gb(0)\|_2\le \frac{\mu_0}{4}.
\end{align*}
\end{lemma}

After deriving sufficient conditions to maintain a positive least eigenvalue at training, we restate the discussion of linear convergence from Section \ref{section:main} formally.

\begin{lemma}\label{lemma:exp}
Consider the linear evolution $\frac{d\fb}{dt} = - \big( \Gb(t) + \frac{\Vb(t)}{\alpha^2}\big)(\fb (t) - \yb)$ from \eqref{eq:flow_evolution}. Suppose that $\lambda_{\min}\big( \Gb(t) + \frac{\Vb(t)}{\alpha^2}\big) \ge \frac{\omega}{2}$ for all times $0 \le t \le T$. Then
\begin{align*}
\|\fb (t)- \yb\|^{2}_{2} \le \exp(-\omega t)\|\fb (0)- \yb\|^2_2 
\end{align*}
for all times $0 \le t \le T$.
\end{lemma}

Using the linear convergence result of Lemma \ref{lemma:exp}, we can now bound the trajectory of the parameters from their initialization.

\begin{lemma}\label{lemma:closew} Suppose that for all $0 \le t \le T$, $\lambda_{\min}\bigg(\Gb(t) + \frac{1}{\alpha^2}\Vb(t)\bigg) \ge \frac{\omega}{2}$ and $|g_k(t) - g_k(0)|  \le R_g \le 1/(m / \delta)^{1 /d}$.
Then with probability $1-\delta$ over the initialization 
\begin{align}
    \|\vb_k(t) - \vb_k(0)\|_2 \le \frac{4\sqrt{n}\|\fb(0)-\yb\|_2}{\alpha \omega \sqrt{m}}\eqqcolon R_v'
\end{align}
for each $k$ and all times $0 \le t \le T$.
\end{lemma}

\begin{lemma}\label{lemma:closeg} Suppose that for all $0 \le t \le T$, $\lambda_{\min}\bigg(\Gb(t) + \frac{1}{\alpha^2}\Vb(t)\bigg) \ge \frac{\omega}{2}$.
Then with probability $1-\delta$ over the initialization 
\begin{align*}|g_k(t) - g_k(0)| \le \frac{4\sqrt{n}\|\fb (0) - \yb \|_2}{\sqrt{m} \omega} \eqqcolon R_g'
\end{align*}
for each $k$ and all times $0 \le t \le T.$
\end{lemma}

The distance of the parameters from initialization depends on the convergence rate (which depends on $\lambda_{\min}(\bLambda(t))$) and the width of the network $m$. We therefore are able to find sufficiently large $m$ for which the maximum parameter trajectories are not too large so that we have that the least eigenvalue of $\bLambda(t)$ is bounded from $0$; this proves the main claim.

Before proving the main results in the case of gradient flow, we use two more technical lemmas. 

\begin{lemma}\label{lemma:init}
Suppose that the network is initialized as \eqref{eq:init} and that $\yb \in \RR^{n}$ has bounded entries $|y_i| \le M$. Then $\|\fb(0) -\yb\|_2 \le C\sqrt{n \log(n / \delta) }$ for some absolute constant $C> 0$.
\end{lemma}

\begin{lemma}[Failure over initialization]\label{lemma:mdelta} Suppose $\vb_1(0), \ldots, \vb_k(0)$ are initialized i.i.d.\ as in \eqref{eq:init} with input dimension $d$. Then with probability $1-\delta$, 
\begin{align*}
\max_{k \in [m] } \frac{1}{\|\vb_k(0)\|_2  } \le \frac{( m / \delta)}{\alpha}^{1 /d} .
\end{align*} 
In addition by \eqref{eq:inc},
for all $t \ge 0$, with probability $1 -\delta$,
\begin{align*} 
\max_{k \in [m] } \frac{1}{\|\vb_k(t)\|_2  } \le \frac{( m / \delta)}{\alpha}^{1 /d}.
\end{align*}
\end{lemma}

\begin{remark*}[Assumption \ref{as:mdelta}]\label{remark:mdelta}
Predominately, machine learning applications reside in the high dimensional regime with $d \ge 50$. Typically $d \gg 50$. This therefore leads to an expression $(m/\delta)^{1/d}$ that is essentially constant. For example, if $d=50$,\ for $\max_{k \in [m] } \frac{1}{\|\vb_k(0)\|_2} \ge 10$, one would need $m / \delta \ge 10^{80}$ (the tail of $\chi^2_d$ also has a factor of $(d/2)!\cdot 2^{d/2}$ which makes the assumption even milder). 
The term $(m/\delta)^{1/d}$ therefore may be taken as a constant for practicality,
\begin{align*}
\max_{k \in [m] } \frac{1}{\|\vb_k(0)\|_2} \le \frac{C}{\alpha}.
\end{align*}
\end{remark*}
While we make Assumption \ref{as:mdelta} when presenting our final bounds, for transparency we do not use Assumption \ref{as:mdelta} during our analysis and apply it only when we present the final over-parametrization results to avoid the overly messy bound.
Without the assumption the theory still holds yet the over-parametrization bound worsens by a power $1+ 1/(d-1)$. This is since the existing bounds can be modified, replacing $m$ with $m^{1 -\frac{1}{d}}$.

\begin{proofTheorem}{\ref{theorem:flow_proof_V}}
By substituting $m =\Omega\big(n^4 \log(n/\delta) / \lambda_0^4  \big)$ and using the bound on $\|\fb(0) -\yb\|_2$ of Lemma \ref{lemma:init}, a direct calculation shows that 
\begin{align*}
    \|\vb_k(t) -\vb_k(0)\|_2 \overset{\text{\ref{lemma:closew}}}{\le} \frac{\alpha \sqrt{n}\|\fb(0) -\yb\|_2}{\sqrt{m} \lambda_0} \le R_v.
\end{align*}
Similarly $m$ ensures that
\begin{align*}
     |g_k(t) -g_k(0)| \overset{\text{\ref{lemma:closeg}}}{\le} \frac{\alpha^2 \sqrt{n}\|\fb(0) -\yb\|_2}{\sqrt{m} \lambda_0} \le R_g.
\end{align*}
The over-parametrization of $m$ implies that the parameter trajectories stay close enough to initialization to satisfy the hypotheses of Lemmas \ref{lemma:rec}, \ref{lemma:Vt} and that $\lambda_{\min}(\bLambda(t)) \ge \lambda_{\min}(\Vb(t))/\alpha^2 \ge \frac{\lambda_0}{2\alpha^2}$.
To prove that $\lambda_{\min}(\bLambda(t)) \ge \frac{\lambda_0}{2\alpha^2}$ holds for all $0 \le t \le T$, we proceed by contradiction and suppose that one of Lemmas \ref{lemma:closew}, \ref{lemma:closeg} does not hold. Take $T_0$ to be the first failure time. Clearly $T_0 >0$ and for $ 0< t < T_0$ the above conditions hold, which implies that $\lambda_{\min}(\Vb(t)) \ge \frac{\lambda_0}{2}$ for $0 \le t \le T_0$; this contradicts one of Lemmas \ref{lemma:closew}, \ref{lemma:closeg} at time $T_0$. Therefore we conclude that Lemmas \ref{lemma:closew}, \ref{lemma:closeg} hold for $t >0$ and we can apply \ref{lemma:exp} to guarantee linear convergence.
\end{proofTheorem}

Here we consider the case where the convergence is dominated by $\Gb$. This occurs when $\alpha > 1$.\\
\begin{proofTheorem}{\ref{theorem:flow_proof_G}}
By substituting $m =\Omega\big(n^4 \log(n/\delta) / \alpha^4 \mu_0^4  \big)$ and using the bound on $\|\fb(0) -\yb\|_2$ of Lemma \ref{lemma:init} we have that 
\begin{align*}
    \|\vb_k(t) -\vb_k(0)\|_2 \overset{\text{\ref{lemma:closew}}}{\le} \frac{ 4 \sqrt{n}\|\fb(0) -\yb\|_2}{\alpha \mu_0 \sqrt{m} } \overset{\text{\ref{lemma:init}}}{\le} \frac{ Cn\sqrt{\log(n /\delta)}}{ \alpha \mu_0 \sqrt{m} } \le  \tilde{R}_v.
\end{align*}
Where the inequality is shown by a direct calculation substituting $m$.

This means that the parameter trajectories stay close enough to satisfy the hypotheses of Lemma \ref{lemma:Gt} if $m = \Omega \big( n^4\log(n/\delta)/\alpha^4 \mu_0^4 \big)$.
Using the same argument as Theorem \ref{theorem:flow_proof_V}, we show that this holds for all $t > 0$. We proceed by contradiction, supposing that one of Lemmas \ref{lemma:closew}, \ref{lemma:closeg} do not hold. Take $T_0$ to be the first time one of the conditions of Lemmas \ref{lemma:closew}, \ref{lemma:closeg} fail. Clearly $T_0 >0$ and for $ 0< t < T_0$ the above derivation holds, which implies that $\lambda_{\min}(\Gb(t)) \ge \frac{\mu_0}{2}$. This contradicts Lemmas \ref{lemma:closew}  \ref{lemma:closeg} at time $T_0$, therefore we conclude that Lemma \ref{lemma:exp} holds for all $t > 0$ and guarantees linear convergence.
\end{proofTheorem}

Note that if $\alpha$ is large, the required complexity on $m$ is reduced. Taking  $\alpha = \Omega(\sqrt{n/\mu_0})$ gives the improved bound

\begin{align*}
    m = \Omega \bigg( \frac{n^2 \log{(n/\delta)}}{\mu_0^2} \bigg).
\end{align*}

\section{Finite Step-size Training}
\label{sec:finite_step}
The general technique of proof for gradient flow extends to finite-step gradient descent. Nonethless, proving convergence for WeightNorm gradient descent exhibits additional complexities arising from the discrete updates and joint training with the new parametrization \eqref{eqn:netclass}. We first introduce some needed notation. 

Define $S_i(R)$ as the set of indices $k \in [m]$ corresponding to neurons that are close to the activity boundary of ReLU at initialization for a data point $\xb_i$, 
\begin{align*}
S_{i}(R) := \{ k \in [m] : \exists~ \vb \text{ with } \|\vb - \vb_k(0)\|_2\le R \text{ and } \ind_{ik}(0) \ne \ind\{\vb^\top \xb_i \ge 0 \} \}.
\end{align*}
We upper bound the cardinality of $|S_i(R)|$ with high probability. 
\begin{lemma}\label{lemma:seminar}
 With probability $1-\delta$, we have that for all $i$
 \begin{align*}
 |S_i(R)| \le \frac{\sqrt{2}mR}{\sqrt{\pi}\alpha} +   \frac{16 \log(n / \delta) }{3}.
 \end{align*}
 \end{lemma}
 Next we review some additional lemmas needed for the proof of Theorems \ref{theorem:finite_step_V}, \ref{theorem:finite_step_G}. Analogous to Lemmas \ref{lemma:closew}, \ref{lemma:closeg}, we bound the finite-step parameter trajectories in Lemmas \ref{lemma:gclose_omega}, \ref{lemma:finite_step_closev}.
\begin{lemma}\label{lemma:gclose_omega}
Suppose the norm of $\|\fb (s)- \yb\|_2^2$ decreases linearly for some convergence rate $\omega$ during gradient descent training for all iteration steps $s=0,1,\ldots, K$ with step-size $\eta$ as $\|\fb (s)- \yb\|_2^2 \le (1 - \frac{\eta \omega}{2})^{s}\| \fb (0)- \yb\|_2^2$ . Then for each $k$ we have
\begin{align*}
|g_k(s) - g_k(0) | \le \frac{4\sqrt{n}\|\fb (0) -  \yb \|_2}{\sqrt{m} \omega}
\end{align*}
for iterations $s=0,1, \ldots, K+1$.
\end{lemma}

\begin{lemma}\label{lemma:finite_step_closev}
Under the assumptions of Lemma \ref{lemma:gclose_omega}, suppose in addition  that $|g_k(s) - g_k(0)| \le 1/(m/ \delta)^{1/d}$ for all iterations steps $s=0,1,  \dots K$ . Then for each $k$,
\begin{align*}
\|\vb_k(s) - \vb_k(0) \|_2\le \frac{8\sqrt{n}\|\fb (0) -  \yb\|_2}{\alpha \sqrt{m} \omega}
\end{align*}
for $s=0,1, \ldots, K+1$.
\end{lemma}

To prove linear rate of convergence
we analyze the $s+1$ iterate error $\|\fb(s+1) - \yb\|_2$ relative to that of the $s$ iterate, $\|\fb(s)- \yb\|_2$. Consider the network's coordinate-wise difference in output between iterations, $f_i(s+1) -f_i(s)$, writing this explicitly based on gradient descent updates yields
\begin{align}\label{eq:finite_diff}
f_i(s+1) - f_i(s) = \frac{1}{\sqrt{m}}\sum_{k =1}^m \frac{c_k g_k(s+1)}{\|\vb_k(s+1)\|_2} \sigma(\vb_k(s+1)^\top \xb_i )  - \frac{c_kg_k(s)}{\|\vb_k(s)\|_2} \sigma(\vb_k(s)^\top \xb_i ).
\end{align}

We now decompose the summand in \eqref{eq:finite_diff} looking at the updates in each layer, $f_i(s+1) -f_i(s) = a_i(s) +b_i(s)$ with \begin{align*}
a_i(s) &= \frac{1}{\sqrt{m}} \sum_{k =1}^m \frac{c_k g_k(s+1)}{\|\vb_k(s+1)\|_2} \sigma(\vb_k(s)^\top \xb_i)  - \frac{c_k g_k(s)}{\|\vb_k(s)\|_2} \sigma(\vb_k (s)^\top \xb_i ), \\
b_i(s) &= \frac{1}{\sqrt{m}}\sum_{k =1}^m \frac{c_k g_k(s+1)}{\|\vb_k(s+1)\|_2} \big( \sigma(\vb_k(s+1)^\top \xb_i )  - \sigma(\vb_k(s)^\top \xb_i ) \big).
\end{align*}
Further, each layer summand is then subdivided into a primary term and a residual. $a_i(s)$, corresponding to the difference in the first layer 
$\bigg(\frac{c_k g_k(s+1)}{\|\vb_k(s+1)\|_2} - \frac{c_k g_k(s)}{\|\vb_k(s)\|_2}\bigg)$, is subdivided into $a_i^{I}(s)$ and $a_i^{II}(s)$ as follows:
\begin{align}\label{eq:A_decomposition}
a_i^{I}(s) &= \frac{1}{\sqrt{m}}\sum_{k =1}^m \bigg(\frac{c_k g_k(s+1)}{\|\vb_k(s)\|_2}- \frac{c_k g_k(s)}{\|\vb_k(s)\|_2}\bigg) \sigma(\vb_k(s)^\top \xb_i), \\
a_i^{II}(s) &= \frac{1}{\sqrt{m}}\sum_{k =1}^m \bigg(\frac{c_k g_k(s+1)}{\|\vb_k(s+1)\|_2}- \frac{c_k g_k(s+1)}{\|\vb_k(s)\|_2}\bigg) \sigma(\vb_k(s)^\top \xb_i). 
\end{align}
 $b_i(s)$ is sub-divided based on the indices in the set $S_i$ that monitor the changes of the rectifiers.  For now, $S_i =S_i(R)$ with $R$ to be set later in the proof. $b_i(s)$ is partitioned to summands in the set $S_i$ and the complement set,
 \begin{align*}
 b_i^{I}(s) &=  \frac{1}{\sqrt{m}}\sum_{k \not\in S_i}  \frac{c_k g_k(s+1)}{\|\vb_k(s+1)\|_2} \big( \sigma(\vb_k(s+1)^\top \xb_i )  - \sigma(\vb_k(s)^\top \xb_i ) \big), \\
 b_i^{II}(s) &=  \frac{1}{\sqrt{m}}\sum_{k \in S_i}  \frac{c_k g_k(s+1)}{\|\vb_k(s+1)\|_2} \big( \sigma(\vb_k(s+1)^\top \xb_i )  - \sigma(\vb_k(s)^\top \xb_i ) \big).
 \end{align*}
 With this sub-division in mind, the terms corresponding to convergence are $\ab^{I}(s), \bbb^{I}(s)$ whereas $\ab^{II}(s), \bbb^{II}(s)$ are residuals that are the result of discretization.
We define the primary and residual vectors $\pb(s), \rb(s)$ as 
\begin{align}\label{eq:residual}
  \pb(s) = \frac{\ab_{I}(s) + \bbb_{I}(s)}{\eta}, \quad \rb(s) =\frac{\ab_{II} + \bbb_{II}(s)}{\eta}.
\end{align}
If the residual $\rb(s)$ is sufficiently small and $\pb(s)$ may be written as $\pb(s) = -\bLambda(s)(\fb(s) - \yb)$ for some iteration dependent evolution matrix $\bLambda(s)$ that has \begin{align}\label{eq:blambda_omega}
    \lambda_{\min}(\bLambda(s)) = \omega/2
\end{align} for $\omega > 0$ then the neural network \eqref{eqn:netclass} converges linearly when trained with WeightNorm gradient descent of step size $\eta$. We formalize the condition on $\rb(s)$ below and later derive the conditions on the over-parametrization ($m$) ensuring that $\rb(s)$ is sufficiently small. 

\begin{property}\label{property:cond}
Given a network from the class \eqref{eqn:netclass} initialized as in \eqref{eq:init} and trained with gradient descent of step-size $\eta$, define the residual $\rb(s)$ as in \eqref{eq:residual} and take $\omega$ as in \eqref{eq:blambda_omega}. We specify the ``residual condition'' at iteration $s$ as
\begin{align*}
\|\rb(s)\|_2 \le c \omega \|\fb(s) -\yb\|_2
\end{align*}
for a sufficiently small constant $c > 0$ independent of the data or initialization.
\end{property}

Here we present Theorem \ref{theorem:finite_step_proof} which is the backbone of Theorems \ref{theorem:finite_step_V} and \ref{theorem:finite_step_G}.
\begin{theorem}\label{theorem:finite_step_proof}
Suppose a network from the class \eqref{eqn:netclass} is trained via WeightNorm gradient descent with an evolution matrix $\bLambda(s)$ as in \eqref{eq:blambda_omega} satisfying $\lambda_{\min}(\bLambda(s)) \ge \omega/2$ for $s=0,1, \ldots K$. In addition if the data meets assumptions \ref{as:parallel},  \ref{as:mdelta}, the step-size $\eta$ of gradient descent satisfies $\eta \le \frac{1}{3 \|\bLambda(s)\|_2}$ and that the residual $\rb(s)$ defined in \eqref{eq:residual} satisfies Property \ref{property:cond} for $s=0,1,\ldots, K$ then we have that
\begin{align*}
\|\fb (s) -\yb\|_2^2 \le \bigg(1 -\frac{\eta \omega}{2} \bigg)^s \|\fb (0) - \yb\|_2^2
\end{align*}
for $s=0, 1, \ldots, K$.
\end{theorem}

\begin{proofTheorem}{\ref{theorem:finite_step_proof}}
This proof provides the foundation for the main theorems. In the proof we also derive key bounds to be used in Theorems \ref{theorem:finite_step_V}, \ref{theorem:finite_step_G}. We use the decomposition we described above and consider again the difference between consecutive terms $\fb (s+1)  -\fb (s)$,
\begin{align}
f_i(s+1) - f_i(s) = \frac{1}{\sqrt{m}}\sum_{k =1}^m \frac{c_k g_k(s+1)}{\|\vb_k(s+1)\|_2} \sigma(\vb_k(s+1)^\top \xb_i )  - \frac{c_kg_k(s)}{\|\vb_k(s)\|_2} \sigma(\vb_k(s)^\top \xb_i ).
\end{align}
Following the decomposition introduced in \eqref{eq:A_decomposition}, $a_i^I(s)$ is re-written in terms of $\Gb(s)$,
\begin{align*}
a_i^I(s) &= \frac{1}{\sqrt{m}}\sum_{k=1}^{m} \frac{c_{k}}{\|\vb_k(s)\|_2} \bigg(-\eta \pdv{L(s)}{g_k}\bigg) \sigma(\vb_k(s)^\top \xb_i) \\
&= -\frac{\eta}{m} \sum_{k=1}^{m} \frac{c_k}{\|\vb_k(s)\|_2} \sum_{j=1}^{n} (f_j(s) - y_j) \frac{c_k}{\|\vb_k(s)\|_2} \sigma(\vb^{\top}_k(s)\xb_j)\sigma(\vb^{\top}_k(s)\xb_i) \\
 &= -\eta \sum_{j=1}^{n}(f_j(s) - y_j)\frac{1}{m} \sum_{k=1}^{m}(c_k)^2 \sigma \bigg(\frac{\vb_k(s)^\top \xb_i}{\|\vb_k(s)\|_2}\bigg) \sigma \bigg(\frac{\vb_k(s)^\top \xb_j}{\|\vb_k(s)\|_2}\bigg) \\
  &= -\eta \sum_{j=1}^{n} (f_j(s) - y_j) \Gb_{ij}(s),
\end{align*}
where the first equality holds by the gradient update rule $g_k(s+1) = g_k(s) - \eta \nabla_{g_k}L(s)$.
In this proof we also derive bounds on the residual terms of the decomposition which we will aid us in the proofs of Theorems \ref{theorem:finite_step_V}, \ref{theorem:finite_step_G}.
$a_i^{I}(s)$ is the primary term of $a_i(s)$, now we bound the residual term $a_i^{II}(s)$.
Recall $a_i^{II}(s)$ is written as
\begin{align*}
    a_i^{II}(s) &= \frac{1}{\sqrt{m}}\sum_{k =1}^m \bigg(\frac{c_k g_k(s+1)}{\|\vb_k(s+1)\|_2}- \frac{c_k g_k(s+1)}{\|\vb_k(s)\|_2}\bigg) \sigma(\vb_k(s)^\top \xb_i),  
\end{align*}
which corresponds to the difference in the normalization in the second layer. Since 
$\nabla_{\vb_k}L(s)$ is orthogonal to $\vb_k(s)$ we have that 
\begin{align*}
&c_k g_k(s+1) \bigg( \frac{1}{\|\vb_k(s+1)\|_2} - \frac{1}{\|\vb_k(s)\|_2} \bigg)\sigma(\vb_k(s)^\top \xb_i)\\
&=c_k g_k(s+1) \bigg( \frac{1}{\sqrt{\|\vb_k(s)\|_2^2+ \eta^2 \|\nabla_{\vb_k}L(s)\|^2_2}} - \frac{1}{\|\vb_k(s)\|_2} \bigg)\sigma(\vb_k(s)^\top \xb_i)\\
&= \frac{-c_kg_k(s+1) \eta^2 \|\nabla_{\vb_k}L(s)\|^2_2}{\|\vb_k(s+1)\|_2\|\vb_k(s)\|_2(\|\vb_k(s)\|_2+ \|\vb_k(s+1)\|_2)} \sigma(\vb_k(s)^\top \xb_i)\\
&\le \frac{-c_k g_k(s+1) \eta^2 \|\nabla_{\vb_k}L(s)\|^2_2}{2\|\vb_k(s)\|_2 \|\vb_k(s+1)\|_2} \sigma\bigg(\frac{\vb_k(s)^\top \xb_i}{\|\vb_k(s)\|_2} \bigg),
\end{align*}
where the first equality above is by completing the square, and the inequality is due to the increasing magnitudes of $\|\vb_k(s)\|_2$.\\

Since $0 \le \sigma \bigg(\frac{\vb_k(s)^\top \xb_i}{\|\vb_k(s)\|_2}\bigg)  \le 1$, the above can be bounded as
\begin{align*}
|a_i^{II}(s)| &\le \frac{1}{\sqrt{m}} \sum_{k=1}^{m} \bigg|\frac{g_k(s+1) \eta^2 \|\nabla_{\vb_k}L(s)\|^2_2}{2\|\vb_k(s)\|_2 \|\vb_k(s+1)\|_2} \bigg| \\
&\le \frac{1}{\sqrt{m}} \sum_{k=1}^{m} \frac{\eta^2 \big(1 + R_g (m/ \delta)^{1/d} \big)^3  n\|\fb (s) -\yb\|_2^2  (m/\delta)^{1 / d}}{\alpha^4 m} \\
& = \frac{\eta^2 n \big(1 +  R_g (m/ \delta)^{1/d}  \big)^3  \|\fb (s) -\yb\|_2^2  (m/\delta)^{1 / d}}{\alpha^4 \sqrt{m}}. \numberthis \label{eq:a_ii}
\end{align*}
The second inequality is the result of applying the bound in equation \eqref{eq:grad_v_bound} on the gradient norm $\|\nabla_{\vb_k}L(s)\|_2$ and using Lemma \ref{lemma:mdelta}.
\par
Next we analyze $b_i(s)$ and sub-divide it based on the sign changes of the rectifiers. Define the set $S_i \coloneqq S_i(R)$ as in Lemma \ref{lemma:seminar} with $R$ taken to be such that $\|\vb_k(s+1) - \vb_k(0)\|_2 \le R$ for all $k$. Take $b_i^{II}(s)$ as the sub-sum of $b_i(s)$ with indices $k$ from the set $S_i$. 

$b_i^{I}(s)$ corresponds to the sub-sum with the remaining indices. By the definition of $S_i$, for $k \not\in S_i$ we have that $\ind_{ik}(s+1) = \ind_{ik}(s)$. This enables us to factor $\ind_{ik}(s)$ and represent $b_i^{I}(s)$ as a Gram matrix similar to $\Vb(s)$ with a correction term from the missing indices in $S_i$.
\begin{align*}
b_i^{I}(s) &=  -\frac{1}{\sqrt{m}} \sum_{k \not\in S_i}  \bigg( \frac{c_k g_k(s+1)}{\|\vb_k(s+1)\|_2} \bigg) \big( \eta \big\langle \nabla_{\vb_k}L(s) ,~\xb_i \big\rangle \big) \ind_{ik}(s)\\
&= -\frac{\eta}{m} \sum_{k \not\in S_i} \bigg( \frac{c_k g_k(s+1)}{\|\vb_k(s+1)\|_2} \bigg) \bigg( \frac{ c_{k}g_{k}(s)}{\|\vb_k(s)\|_2} \bigg) \sum_{j =1}^{n}  (f_j(s) - y_j) \ind_{ik}(s) \ind_{jk}(s) \big\langle \xb_j^{\vb_k(s)^\perp},~  \xb_i \big \rangle .
\end{align*}

Note that $\big \langle \xb_j^{\vb_k (s) ^\perp},~ \xb_i \big \rangle= \big \langle \xb_j^{\vb_k(s)^\perp},~ \xb_i^{\vb_k(s)^\perp} \big\rangle$ therefore,
\begin{align*}
b_i^{I}(s) &=  -\frac{\eta}{m} \sum_{k \not\in S_i} \bigg( \frac{c_k g_k(s+1)}{\|\vb_k(s+1)\|_2} \bigg) \bigg( \frac{ c_{k}g_{k}(s)}{\|\vb_k(s)\|_2} \bigg) \sum_{j =1}^{n}  (f_j(s) - y_j) \ind_{ik}(s) \ind_{jk}(s) \big \langle \xb_j^{\vb_k(s)^\perp},~  \xb_i^{\vb_k(s)^\perp} \big \rangle.
\end{align*}

Define $\tilde{\Vb}(s)$ as 
\begin{align*}
\tilde{\Vb}_{ij}(s) = \frac{1}{m} \sum_{k=1}^{m} \bigg( \frac{\alpha c_k g_k(s+1)}{\|\vb_k(s+1)\|_2} \bigg) \bigg( \frac{ \alpha c_{k}g_{k}(s)}{\|\vb_k(s)\|_2} \bigg)  \ind_{jk}(s) \ind_{ik}(s) \big \langle \xb_i^{\vb_k(s)^\perp},~  \xb_j^{\vb_k(s)^\perp} \big \rangle.
\end{align*}
This matrix is identical to $\Vb(s)$ except for a modified scaling term $\big( \frac{c_k^2g_k(s+1)g_k(s)}{\|\vb_k(s)\|_2 \|\vb_k(s+1)\|_2}\big)$. We note however that
\begin{align*}
    \min \Bigg( \bigg(\frac{c_k g_k(s)}{\|\vb_k(s)\|_2} \bigg)^2, \bigg( \frac{c_k g_k(s+1)}{\|\vb_k(s+1)\|_2}\bigg)^2 \Bigg) &\le \bigg( \frac{c_k g_k(s)}{\|\vb_k(s)\|_2} \bigg) \bigg( \frac{c_k g_k(s+1)}{\|\vb_k(s+1)\|_2} \bigg)\\
    &\le \max \Bigg( \bigg(\frac{c_k g_k(s)}{\|\vb_k(s)\|_2} \bigg)^2, \bigg( \frac{c_k g_k(s+1)}{\|\vb_k(s+1)\|_2}\bigg)^2 \Bigg)
\end{align*}
because $g_k(s), c_k^2$ are positive.
Hence the matrix $\tilde{\Vb}(s)$ satisfies the hypothesis of Lemma \ref{lemma:ht} entirely.
We write $b_i^{I}(s)$ as
\begin{align*}
b_i^{I}(s) = -\eta/ \alpha^2 \sum_{j=1}^{n} (f_j(s) - y_j) (\tilde{\Vb}_{ij}(s) - \tilde{\Vb}_{ij}^\perp (s)) ,
\end{align*}
where we have defined 
\begin{align}\label{eq:define_vperp}
\tilde{\Vb}_{ij}^\perp(s) = \frac{1}{m} \sum_{k \in S_i} \bigg(\frac{\alpha c_kg_k(s)}{\|\vb_k(s)\|_2}\bigg)\bigg(\frac{\alpha c_kg_k(s+1)}{\|\vb_k(s+1)\|_2} \bigg) \ind_{ik}(s)\ind_{jk}(s) \big \langle \xb_i^{\vb_k(s)^\perp},~  \xb_j^{\vb_k(s)^\perp}\big \rangle.
\end{align}

We then bound the magnitude of each entry $\tilde{\Vb}_{ij}^{\perp}(s)$:
\begin{align*}
\tilde{\Vb}_{ij}^{\perp}(s) &= \frac{1}{m} \sum_{k \in S_i} \bigg(\frac{\alpha c_kg_k(s)}{\|\vb_k(s)\|_2} \bigg)\bigg(\frac{\alpha c_kg_k(s+1)}{\|\vb_k(s+1)\|_2} \bigg) \ind_{ik}(s)\ind_{jk}(s) \big \langle \xb_i^{\vb_k(s)^\perp},~  \xb_j^{\vb_k(s)^\perp} \big \rangle \\
& \le \frac{(1+ R_{g}(m /\delta)^{1 /d})^2|S_{i}|}{ m }. \numberthis \label{eq:vperp_ij}
\end{align*}
Lastly we bound the size of the residual term $b_i^{II}(s)$,
\begin{align*}
    |b_{i}^{II}(s)| &= \bigg|-\frac{1}{\sqrt{m}} \sum_{k \in S_i} \frac{c_k g_k(s+1)}{\|\vb_k(s+1)\|_2}\bigg(\sigma(\vb_k(s+1)^\top \xb_i) - \sigma(\vb_k(s)^{\top}\xb_i)\bigg) \bigg| \\
    &\le \frac{g_k (s+1) \eta |S_i| \cdot \|\nabla_{\vb_k}L(s)\|_2}{\sqrt{m}\|\vb_k(s+1)\|_2} \\
    &\le \frac{\eta  |S_i| (1 + (m/\delta)^{1/d}R_g) \|\nabla_{\vb_k} L(s) \|_2}{\alpha\sqrt{m}}.
\end{align*}
Where we used the Lipschitz continuity of $\sigma$ in the first bound, and took $R_g>0$ that satisfies $|g_k(s+1) -g_k(0)| \le R_g$ in the second inequality. Applying the bound \eqref{eq:grad_v_bound},
\begin{align}\label{eq:b_i_II_bound}
    |b_i^{II}(s)| \le \frac{\eta|S_i| \sqrt{n} (1+ R_g(m/\delta)^{1/d})^2 \|\fb(s) -\yb\|_2}{\alpha^2 m}.
\end{align}

The sum $\fb(s+1) - \fb(s) = \ab^{I}(s) + \ab^{II}(s) + \bbb^{I}(s) + \bbb^{II}(s)$ is separated into the primary term $\eta \pb(s) = \ab_{I}(s) +\bbb_{I}(s)$ and the residual term $\eta \rb(s) = \ab_{II}(s) + \bbb_{II}(s)$ which is a result of the discretization. With this, the evolution matrix $\bLambda(s)$ in \eqref{eq:blambda_omega} is re-defined as
\begin{align*}
    \bLambda(s) \coloneqq \Gb(s) + \frac{\tilde{\Vb}(s) -\tilde{\Vb}^\perp(s)}{\alpha^2}
\end{align*}
and 
\begin{align*}
    \fb(s+1) - \fb(s) = -\eta\bLambda(s)(\fb(s) - \yb) +\eta \rb(s).
\end{align*}
Now we compare $\|\fb (s+1) - \yb\|_2^2$ with $\|\fb (s) - \yb\|_2^2$,
\begin{align*}
\|\fb (s+1) - \yb\|_2^2  =& \|\fb (s+1) -\fb (s) +\fb (s) - \yb\|_2^2 \\
=&\|\fb (s) - \yb\|_2^2 + 2\big \langle \fb (s+1) - \fb (s),~\fb (s) - \yb\big \rangle \\
&+\big \langle \fb (s+1)- \fb (s),~\fb (s+1) - \fb (s) \big \rangle.\\
\end{align*}
Substituting  
\begin{align*}
    \fb(s+1) - \fb(s) = \ab^{I}(s) + \bbb^{I}(s) +\ab^{II}(s)+ \bbb^{II}(s) = -\eta \bLambda(s)(\fb(s) - \yb) + \eta\rb(s)
\end{align*}
we obtain
\begin{align*}
\|\fb (s+1) - \yb\|_2^2 =& \|\fb (s) -\yb\|_2^2 + 2(-\eta\bLambda(s)(\fb (s) - \yb) + \eta \rb(s)) ^\top (\fb (s) - \yb) \\
&+\eta^2(\bLambda(s)(\fb (s) - \yb) - \rb(s))^\top (\bLambda(s)(\fb (s) - \yb) - \rb(s)) \\
\le& \|\fb (s) - \yb\|_2^2 + (\fb (s) - \yb)^\top( -\eta \bLambda(s) + \eta^2 \bLambda^2(s))(\fb (s) - \yb)\\
&+ \eta \rb(s)^{\top}(\Ib - \eta \bLambda(s))(\fb (s) - \yb) + \eta^2 \|\rb(s)\|_2^2.
\end{align*}
Now as $\lambda_{\min}(\bLambda(s)) \ge \omega/2$ and $\eta = \frac{1}{3\|\bLambda(s) \|_2}$, we have that
\begin{align*}
(\fb (s) - \yb)^\top( -\eta \bLambda(s) + \eta^2 \bLambda^2(s))(\fb (s) - \yb) = -\eta(\fb (s) -\yb)^\top(\Ib - \eta \bLambda(s))\bLambda(s)(\fb (s) - \yb) \le -\frac{3\eta\omega}{8}\|\fb (s)  - \yb\|_2^2.
\end{align*}
Next we analyze the rest of the terms and group them as $\qb(s)$,
\begin{align*}
\qb(s) &\coloneqq \eta \rb(s)^\top (\Ib - \eta \bLambda(s)) (\fb(s) -\yb) + \eta^2\|\rb(s)\|_2^2 \\
&\le  \eta \|\rb(s)\|_2\|\fb(s) -\yb\|_2 + \eta ^2\|\rb(s)\|_2^2 .%\\
\end{align*}
By Property~\ref{property:cond} we have 
\begin{align*}
\qb(s) \le \eta c \omega \|\fb(s) - \yb\|_2^2( 1 + \eta c\omega) \le 2c \eta \omega \|\fb(s)- \yb\|_2^2 , 
\end{align*}
so that 
\begin{align*}
\qb(s) \le c' \eta \omega \|\fb (s) -\yb\|_2^2, 
\end{align*}
for $c'$ sufficiently small.  Substituting, the difference $\fb(s+1) -\yb$ is bounded as
\begin{align*}
\|\fb (s+1) - \yb\|_2^2  &\le \|\fb (s) - \yb\|_2^2 -\eta\omega(1- \eta  \|\bLambda(s)\|_2)\|\fb (s)  - \yb\|_2^2 + c' \eta \omega \|\fb (s) -\yb\|_2^2 \\
&\le (1 - \eta \omega(1 - \eta \|\bLambda(s)\|_2) + c' \eta \omega)\|\fb(s) - \yb\|_2^2 \\
&\le (1 - \eta\omega /2)\|\fb (s) - \yb\|_2^2, 
\end{align*}
for well chosen absolute constant $c$. Hence for each $s =0,1, \ldots, K$,
\begin{align*}
    \|\fb(s+1) -\yb\|_2^2 \le (1-\eta\omega/2)\|\fb(s) -\yb\|_2^2, 
\end{align*}
 so the prediction error converges linearly.
\end{proofTheorem}

 In what comes next we prove the necessary conditions for Property \ref{property:cond}, and define the appropriate $\omega$ for the $\Vb$ and $\Gb$ dominated regimes, in order to show $\lambda_{\min}(\bLambda(s)) \ge \omega/2$.

\begin{proofTheorem}{\ref{theorem:finite_step_V}}
To prove convergence we would like to apply Theorem \ref{theorem:finite_step_proof} with $\omega/2 = \frac{\lambda_0}{2\alpha^2}$. To do so we need to show that $m =\Omega\big(n^4 \log(n/ \delta) /\lambda_0^4 \big)$ guarantees that Property \ref{property:cond} holds and that $\lambda_{\min}(\bLambda(s)) \ge \lambda_0 / 2\alpha^2$. For finite-step gradient training, take 
\begin{align}\label{eq:Gu0001}
R_v = \frac{\alpha \lambda_0  }{192n
(m/ \delta)^{1 /d}}, \quad R_g = \frac{\lambda_0}{96n(m/ \delta)^{1 /d}} .
\end{align}
Note the residual $\rb(s)$ and the other terms $\bbb_{I}(s), \bbb_{II}(s)$ depend on the sets $S_i$ that we define here using $R_v$.
We make the assumption that $\|\vb_k(s) - \vb_k(0) \|_2\le R_v$ and $ |g_k(s) - g_k(0) |  \le R_g$ for all $k$ and that $s =0,1, \ldots K+1$, this guarantees that $\bbb_{I}(s)$ and $\bLambda(s)$ are well defined. Applying Lemmas \ref{lemma:hzero}, \ref{lemma:Vt} with $R_v, R_g$ defined above, we have that $\lambda_{\min}(\tilde{\Vb}(s)) \ge \frac{5\lambda_0}{8}$.
Then the least eigenvalue of the evolution matrix $\bLambda(s)$ is bounded below
\begin{align*}
    \lambda_{\min}(\bLambda(s)) &= \lambda_{\min}\bigg( \Gb(s) + \frac{\tilde{\Vb}(s) - \tilde{\Vb}^{\perp}(s)}{\alpha^2} \bigg) \\
    &\ge \lambda_{\min}\bigg( \frac{\tilde{\Vb}(s) - \tilde{\Vb}^{\perp}(s)}{\alpha^2} \bigg) \\
    &= \frac{\lambda_{\min} (\tilde{\Vb}(s) - \tilde{\Vb}^{\perp}(s))
    }{\alpha^2} \\ 
    & \ge \frac{5\lambda_0}{8\alpha^2} -\frac{\|\tilde{\Vb}^\perp(s)\|_2}{\alpha^2}. 
\end{align*}

The first inequality holds since $\Gb(s) \succ 0$ and the last inequality is since $\lambda_{\min}(\tilde{\Vb}(s)) \ge \frac{5\lambda_0}{8}$.

To show $\lambda_{\min}(\bLambda(s)) \ge \frac{\lambda_0}{2\alpha^2}$ we bound $\|\tilde{\Vb}^\perp(s)\|_2 \le \frac{\lambda_0}{8}$.
By \eqref{eq:vperp_ij}, we have 
\begin{align*}
    |\tilde{\Vb}_{ij}^{\perp}(s)| \le \frac{(1 + R_g(m/\delta)^{1/d})|S_i|}{m} \le (1+ R_g(m/\delta)^{1/d})\bigg(\frac{\sqrt{2}\tilde{R_v}}{\sqrt{\pi}\alpha} + \frac{16\log(n/\delta)}{3m} \bigg).
\end{align*}
Substituting $R_v, R_g$ and $m$, a direct calculation shows that
\begin{align*}
    |\tilde{\Vb}_{ij}^{\perp}(s)| \le \frac{\lambda_0}{8n},
\end{align*}
which yields
\begin{align*}
    \|\tilde{\Vb}^{\perp}(s)\|_2  \le \|\tilde{\Vb}^{\perp}(s)\|_{F} \le \frac{\lambda_0}{8}.
\end{align*}

Hence $\lambda_{\min}(\bLambda(s)) \ge \frac{\lambda_0}{2\alpha^2}$ for iterations $s=0,1, \ldots K$.

We proceed by showing the residual $\rb(s)$ satisfies property \ref{property:cond}.
Recall $\rb(s)$ is written as 
\begin{align*}
   \rb(s) = \frac{\ab^{II}(s)}{\eta}+ \frac{\bbb^{II}(s)}{\eta}.
\end{align*}
and Property \ref{property:cond} states that $\|\rb(s)\|_2 \le \frac{c \eta \lambda_0}{\alpha^2} \|\fb(s) - \yb\|_2$ for sufficiently small absolute constant $c < 1$.
This is equivalent to showing that both $\ab^{II}(s)$, $\bbb^{II}(s)$ satisfy
\begin{align}\label{eq:a2b2_condition}
    \|\ab^{II}(s)\|_2&\le  \frac{c \eta \lambda_0}{\alpha^2} \|\fb(s) -\yb\|_2,\\
    \|\bbb^{II}(s)\|_2&\le  \frac{c \eta \lambda_0}{\alpha^2} \|\fb(s) -\yb\|_2.
\end{align}
We consider each term at turn. By \eqref{eq:b_i_II_bound},
\begin{align*}
\|\bbb^{II}(s)\|_2&\le \sqrt{n}\max_{i}b_i^{II}(s) \\
&\le \max_{i}\frac{\eta n (1 + R_g(m /\delta)^{1/d})^2|S_i|\|\fb(s)-\yb\|_2}{\alpha^2 m}\\
&\le \frac{CmR_{v}\eta n\|\fb(s)-\yb\|_2}{\alpha^2 m} \\
&\le \frac{\lambda_0 \eta\|\fb(s) -\yb\|_2}{\alpha^2} \cdot nCR_v.\\
\end{align*}
In the above we used the values of $R_v, R_g$ defined in \eqref{eq:Gu0001} and applied Lemma \ref{lemma:seminar} in the third inequality.
Taking $m =\Omega\big(n^4 \log(n/ \delta) /\lambda_0^4 \big)$ with large enough constant yields
\begin{align*}
    \|\bbb^{II}(s)\|_2\le \frac{c\lambda_0 \eta\|\fb(s) -\yb\|_2}{\alpha^2}.
\end{align*}
Next we analogously bound $\|\ab^{II}(s)\|$ via the bound \eqref{eq:a_ii},
\begin{align*}
   \| \ab^{II}(s)\|_2&\le \sqrt{n}\max_{i}a^{II}_i(s)\\
   &\le   \frac{\eta^2 n^{3/2} \big(1 +  R_g (m/ \delta)^{1/d}  \big)^3  \|\fb (s) -\yb\|_2^2  (m/\delta)^{1 / d}}{\alpha^4 \sqrt{m}}\\
   &\le \frac{\eta \lambda_0 \|\fb(s) - \yb\|_2}{\alpha^2}\cdot \frac{\eta \big(1 +  R_g (m/ \delta)^{1/d}  \big)^3 n^{3/2} \|\fb (s) -\yb\|_2(m/\delta)^{1 / d}}{\lambda_0\alpha^2\sqrt{m}} \\
   & \le \frac{\eta \lambda_0 \|\fb(s) - \yb\|_2}{\alpha^2}\cdot \frac{\eta}{ \alpha^2}\cdot\frac{Cn^2\sqrt{\log(n/\delta)}}{\lambda_0\sqrt{m}}\\
   &\le c\eta \omega \|\fb(s) -\yb\|_2.
\end{align*}
In the above we applied Lemma \ref{lemma:init} once again. The last inequality holds since $m = \Omega(n^4\log(n/ \delta) /\lambda_0^4)$ and $\eta = O\bigg(\frac{\alpha^2}{\|\Vb(s)\|_2}\bigg)$, hence $\rb(s)$ satisfies Property \ref{property:cond}.
Now since Theorem \ref{theorem:finite_step_proof} holds with $\omega = \lambda_0/\alpha^2$ we have that the maximum parameter trajectories are bounded as $\|\vb_k(s) - \vb_k(0)\|_2 \le R_v$ and $\|g_k(s) -g_k(0)\| \le R_g$ for all $k$ and every iteration $s = 0, 1, \ldots, K+1$ via Lemmas \ref{lemma:gclose_omega}, \ref{lemma:finite_step_closev}.

To finish the proof, we apply the same contradiction argument as in Theorems \ref{theorem:flow_proof_V}, \ref{theorem:flow_proof_G}, taking the first iteration $s = K_0$ where one of Lemmas \ref{lemma:gclose_omega}, \ref{lemma:finite_step_closev} does not hold. We note that $K_0 > 0$ and by the definition of $K_0$, for $s=0,1, \ldots, K_0-1$ the Lemmas \ref{lemma:gclose_omega}, \ref{lemma:finite_step_closev} hold which implies that by the argument above we reach linear convergence in iteration $s =K_0$. This contradicts one of Lemmas \ref{lemma:gclose_omega}, \ref{lemma:finite_step_closev} which gives the desired contradiction, so we conclude that we have linear convergence with rate $\lambda_0/2\alpha^2$ for all iterations.
\end{proofTheorem}

\begin{proofTheorem}{\ref{theorem:finite_step_G}}
For $\Gb$-dominated convergence, we follow the same steps as in the proof of Theorem \ref{theorem:finite_step_V}. 
We redefine the trajectory constants for the finite step case
\begin{align*}
    \tilde{R}_v \coloneqq\frac{\sqrt{2\pi} \alpha \mu_0 }{64 n (m /\delta)^{1/d}},\quad R_g  \coloneqq \frac{\mu_0}{48n(m/ \delta)^{1 /d}}.
\end{align*}

To use Theorem \ref{theorem:finite_step_proof} we need to show that $m =\Omega\big(n^4 \log(n/ \delta) / \alpha^4  \mu_0^4 \big)$ guarantees Property \ref{property:cond}, and that $\lambda_{\min}(\bLambda(s)) \ge \mu_0 / 2$. 
We again note that the residual $\rb(s)$ and  $\bbb_{I}(s), \bbb_{II}(s)$ depend on the sets $S_i$ that we define here using $\tilde{R}_v$ above as $S_i \coloneqq  S_i(\tilde{R}_v)$.

We start by showing the property on the least eigenvalue. We make the assumption that we have linear convergence with $\omega/2 = \mu_0/2$ and step-size $\eta$ for iterations $s=0, \ldots K$ so that Lemmas \ref{lemma:gclose_omega}, \ref{lemma:finite_step_closev} hold. 
Via an analogous analysis of the continous case we reach that $m =\Omega\big(n^4 \log(n/ \delta) / \mu_0^4\alpha^4  \big)$ implies
\begin{align*}
  \|\vb_k(s) - \vb_k(0) \|_2\le \frac{16\alpha\sqrt{n}\|\fb (0) -  \yb\|_2}{\alpha \sqrt{m} \mu_0} \le \tilde{R}_v,   \quad    |g_k(s) - g_k(0) | \le \frac{8\sqrt{n}\|\fb (0) -  \yb \|_2}{\sqrt{m} \mu_0} \le R_g.
\end{align*}
for $s=0, \ldots K+1$
by Lemmas \ref{lemma:gclose_omega}, \ref{lemma:finite_step_closev} and that $\bLambda(s), \bbb_{I}(s)$ are well defined.
Using the bounds on the parameter trajectories, Lemma \ref{lemma:Gt} and $\tilde{R}_v$ defined above yield $\lambda_{\min}(\Gb(s)) \ge \frac{5\mu_0}{8}.$
The least eigenvalue of the evolution matrix $\bLambda(s)$ is bounded below as
\begin{align*}
    \lambda_{\min}(\bLambda(s)) &= \lambda_{\min}\bigg( \Gb(s) + \frac{\tilde{\Vb}(s) - \tilde{\Vb}^{\perp}(s)}{\alpha^2} \bigg) \\
    &\ge \lambda_{\min}(\Gb(s)) - \|\tilde{\Vb}^{\perp}(s)\|_2
\end{align*}
since $\tilde{\Vb}(s) \succ 0$ and $\alpha \ge 1$.
We bound the spectral norm of $\|\tilde{\Vb}^{\perp}(s)\|_2$, for each entry $i,j$ we have by \eqref{eq:vperp_ij} that
\begin{align*}
    |\tilde{\Vb}_{ij}^{\perp}(s)| &\le \frac{(1 + R_g(m/\delta)^{1/d})|S_i|}{m} \\
    &\le (1+ R_g(m/\delta)^{1/d})\bigg(\frac{\sqrt{2}\tilde{R_v}}{\sqrt{\pi}\alpha} + \frac{16\log(n/\delta)}{3m} \bigg)  \\
    &\le \frac{8 \tilde{R}_v}{\sqrt{2 \pi}\alpha} \\
    &\le \frac{\mu_0}{8n}.
\end{align*}
where in the above inequalities we used our bounds on $\tilde{R}_v, R_g$ and $m$. Then the spectral norm is bounded as 
\begin{align*}
    \|\tilde{\Vb}^{\perp}(s)\|_2  \le \|\tilde{\Vb}^{\perp}(s)\|_{F} \le \mu_0/8.
\end{align*}
Hence we have that $\lambda_{\min}(\bLambda(s)) \ge \mu_0/2$ for  $s = 0,1, \dots K$.

Next we show the residual $\rb(s)$ satisfies Property \ref{property:cond}.
Recall $\rb(s)$ is written as 
\begin{align*}
   \rb(s)= \frac{\ab^{II}(s)}{\eta}+ \frac{\bbb^{II}(s)}{\eta}.
\end{align*}
Property \ref{property:cond} states the condition $\|\rb(s)\|_2 \le c\omega\eta \|\fb(s) - \yb\|_2$ for sufficiently small $c < 1$ with $\omega =\mu_0$.
This is equivalent to showing that both $\ab^{II}(s)$, $\bbb^{II}(s)$ satisfy that
\begin{align}
    \|\ab^{II}(s)\|_2&\le c\eta \mu_0  \|\fb(s) -\yb\|_2,\\
    \|\bbb^{II}(s)\|_2&\le c\eta \mu_0 \|\fb(s) -\yb\|_2, 
\end{align}
for sufficiently small absolute constant $c$.
For $\bbb_{II}(s)$ we have that \eqref{eq:b_i_II_bound} gives
\begin{align*}
\|\bbb^{II}(s)\|_2&\le \sqrt{n}\max_{i}b_i^{II}(s) \\
&\le \max_{i}\frac{\eta(1 + R_g(m /\delta)^{1/d})^2|S_i|n\|\fb(s)-\yb\|_2}{\alpha^2 m}.
\end{align*}

Next applying Lemmas \ref{lemma:seminar} and \ref{lemma:init} in turn yields
\begin{align*}
&\le \frac{Cm \tilde{R}_{v}\eta n\|\fb(s)-\yb\|_2}{\alpha^2 m} \\
&\le \eta \mu_0 \|\fb(s) -\yb\|_2 \frac{\tilde{R}_v}{n\alpha^2 }. 
\end{align*}
Substituting $m =\Omega\big(n^4 \log(n/ \delta) / \mu_0^4\alpha^4  \big)$ for a large enough constant and $R_v$
we get \begin{align*}
    \|\bbb^{II}(s)\|_2\le c \eta \mu_0 \|\fb(s) -\yb\|_2.
\end{align*}
Analogously we bound $\|\ab^{II}(s)\|_2$ using \eqref{eq:a_ii},
\begin{align*}
   \| \ab^{II}(s)\|_2&\le \sqrt{n}\max_{i}a_i^{II}(s)\\ &\le   \frac{\eta^2 n^{3/2} \big(1 +  R_g (m/ \delta)^{1/d}  \big)^3  \|\fb (s) -\yb\|_2^2  (m/\delta)^{1 / d}}{\alpha^4 \sqrt{m}}\\
   &\le \eta \mu_0 \|\fb(s) - \yb\|_2 \cdot \frac{\eta \big(1 +  R_g (m/ \delta)^{1/d}  \big)^3 n^{3/2} \|\fb (s) -\yb \|_2(m/\delta)^{1 / d}}{\mu_0\alpha^4\sqrt{m}} \\
   & \le \eta \mu_0 \|\fb(s) - \yb\|_2 \cdot \frac{\eta}{\alpha^2}\cdot\frac{ Cn^2\sqrt{\log(n/\delta)}}{\alpha^2\mu_0^2 \sqrt{m}}\\
   & \le c\eta \mu_0 \|\fb(s) -\yb\|_2.
\end{align*}
Where we have used Lemma \ref{lemma:init} in the third inequality and substituted $m = \Omega(n^4\log(n/ \delta) /\alpha^4\mu_0^4)$ noting that $\eta = O\big(\frac{1}{\|\bLambda(s)\|_2}\big)$ and that $\alpha \ge 1$ in the last inequality.
Therefore we have that $\rb(s)$ satisfies Property \ref{property:cond} so that Theorem \ref{theorem:finite_step_proof} holds. By the same contradiction argument as in Theorem \ref{theorem:finite_step_V} we have that this holds for all iterations.
\end{proofTheorem}
\section{Additional Technical Lemmas and Proofs of the Lemmas from Appendix~\ref{appendix:flow_proof}}\label{appendix:flow_lemmas}

\begin{proofLemma}{\ref{lemma:hau}}
We prove Lemma \ref{lemma:hau} for $\Vb^{\infty}$, $\Gb^{\infty}$ separately.
 $\Vb^{\infty}$ can be viewed as the covariance matrix of the functionals $\phi_i$ defined as
\begin{align}\label{eq:phis}
\phi_i(\vb) = \xb_i\bigg(\Ib- \frac{\vb\vb^{\top}}{\|\vb\|^2_2}\bigg)\ind \{ \vb^\top \xb_i \ge 0\}
\end{align}
over the Hilbert space $\mathcal{V}$ of $L^{2}(N(0,\alpha^2 \Ib))$ of functionals. Under this formulation, if $\phi_1, \phi_2, \ldots, \phi_n$ are linearly independent, then $\Vb^{\infty}$ is strictly positive definite. Thus, to show that $\Vb^{\infty}$ is strictly positive definite is equivalent to showing that
\begin{equation}\label{eq:cV}
    c_1 \phi_1 + c_2 \phi_2 + \cdots + c_n \phi_n = 0\;\; \text{in } \mathcal{V} 
\end{equation}
implies $c_i = 0$ for each $i$. The $\phi_i$s are piece-wise continuous functionals, and equality in $\cV$ is equivalent to 
\begin{align*}
c_1 \phi_1 + c_2 \phi_2 + \cdots + c_n \phi_n = 0 \;\; \text{almost everywhere} . 
\end{align*}
For the sake of contradiction, assume that there exist $c_1,\ldots, c_n$ that are not identically $0$, satisfying \eqref{eq:cV}. As $c_i$ are not identically $0$, there exists an $i$ such that $c_i \ne 0$. We show this leads to a contradiction by constructing a non-zero measure region such that the linear combination $\sum_{i} c_i \phi_i$ is non-zero.\par 
Denote the orthogonal subspace to $\xb_i$ as $D_i \coloneqq \{ \vb \in \RR^d: \vb^{\top}\xb_i = 0 \}$. By Assumption~\ref{as:parallel}, 
\begin{align*}
    D_i \not\subseteq \bigcup_{j \ne i} D_j %D_k.. 
\end{align*} 
This holds since $D_i$ is a $(d-1)$-dimensional space which may not be written as the finite union of sub-spaces $D_i \cap D_j$ of dimension $d-2$ (since $\xb_i$ and $\xb_j$ are not parallel). 
Thus, take $\zb \in D_i \backslash \bigcup_{j \ne i} D_j$. Since  
$\bigcup_{j \ne i} D_j$ is closed in $\RR^{d}$, there exists an $R > 0$ such that
\begin{align*}
\cB(\zb, 4R) \cap \bigcup_{j \ne i} D_j = \emptyset. 
\end{align*}
Next take $\yb \in  \partial \cB(\zb, 3R) \cap D_i$ 
(where $\partial$ denotes the boundary)
on the smaller disk of radius $3R$ so that it satisfies  $\|\yb\|_2= \max_{\yb' \in \partial \cB(\zb, 3R) \cap D_i} \|\yb'\|_2$.  Now for any $r \le R$,
the ball $\mathcal{B}(\yb, r)$ is such that for all points $\vb \in \mathcal{B}(\yb, r)$ we have $\|\vb^{\xb_i^\perp} \|_2\ge 2R $ and $\|\vb^{\xb_i} \|_2\le R$.
Then for any $r \le R$, the points $\vb \in \cB(\yb,r) \subset \cB(\zb,4R)$ satisfy that
\begin{align*}
\|\xb_i^{\vb^\perp}\|_2\ge \|\xb_i\|_2- \frac{\xb_i \cdot \vb}{\|\vb\|_2} \ge \|\xb_i\|_2\bigg(1  - \frac{R}{2R} \bigg) \ge \frac{\|\xb_i\|_2}{2}. \end{align*}

Next we decompose the chosen ball $\cB(\yb,r) = B^{+}(r)  \vee B^{-}(r)$ to the areas where the ReLU function at the point $\xb_i$ is active and inactive
\begin{align*}
B^+(r) = \cB(\yb,r) \cap \{\xb_i ^\top \vb  \ge 0 \} ,\quad B^-(r) = \cB(\yb,r) \cap \{\xb_i^\top \vb < 0 \}.
\end{align*}
Note that $\phi_i$ has a discontinuity on $D_i$ and is continuous within each region $B^+(r)$ and $B^-(r)$. Moreover, for $j \ne i$, $\phi_j$ is continuous on the entire region of $\cB(\yb,r)$ since $\cB(\yb,r) \subset \cB(\zb, 4R) \subset D_j^{c}$. Since we have that $\phi_j$ is continuous in the region, the Lebesgue differentiation theorem implies that for $r \rightarrow 0$, $\phi_i$ satisfies on $B^+(r), B^-(r)$:
\begin{align*}
\lim_{r \rightarrow 0} \frac{1}{\mu(B^+(r))}\int_{B^{+}(r)}\phi_i  = \xb_i^{\yb^{\perp}} \ne 0, \quad 
\lim_{r \rightarrow 0} \frac{1}{\mu(B^-(r))}\int_{B^{-}(r)}\phi_i  = 0.
\end{align*}

For $j \ne i$  $\phi_j$ is continuous on the entire ball $\cB(\yb,r)$ hence the Lebesgue differentiation theorem also gives
\begin{align*}
\lim_{r \rightarrow 0} \frac{1}{\mu(B^+(r))}\int_{B^{+}(r)}\phi_i  = \phi_j(\yb),\quad
\lim_{r \rightarrow 0} \frac{1}{\mu(B^-(r))}\int_{B^{-}(r)}\phi_i  = \phi_j(\yb).
\end{align*}

We integrate $c_1 \phi_1 + \dots c_n \phi_n$  over $B^-(r)$ and $B^+(r)$ separately and subtract the integrals. By the assumption, $c_1 \phi_1 + \cdots + c_n \phi_n=0$ almost everywhere so each integral evaluates to $0$ and the difference is also $0$,
\begin{align}\label{eq:diff_comb}
0= \frac{1}{\mu(B^{+}(r))} \int_{B^+(r)} c_1 \phi_1 + \cdots + c_n \phi_n - \frac{1}{\mu(B^{-}(r))}\int_{B^-(r)} c_1 \phi_1 + \cdots + c_n \phi_n .
\end{align}
By the continuity of $\phi_j$ for $j \ne i$ taking $r \rightarrow 0$ we have that
\begin{align*}
\frac{1}{\mu(B^{+}(r))} \lim_{r\rightarrow 0} \int_{B^+(r)} \phi_j - \frac{1}{\mu(B^{-}(r))}\int_{B^-(r)}\phi_j &= 
\phi_j(\yb) - \phi_j(\yb) = 0.
\end{align*} 
For $\phi_i$ the functionals evaluate differently. For $B^-(r)$ we have that
\begin{align*}
\frac{1}{\mu(B^{-}(r))}\lim_{r\rightarrow 0}\int_{B^-(r)} \phi_i = 
\frac{1}{\mu(B^{-}(r))}\lim_{r\rightarrow 0}\int_{B^-(r)}  0 = 0,\\
\end{align*}
while the integral over the positive side, $B^{+}(r)$ is equal to
\begin{align*}
\frac{1}{\mu(B^{+}(r))}\int_{B^+(r)} \phi_i(\zb) d\zb &= 
\frac{1}{\mu(B^{+}(r))}\int_{B^+(r)} \xb_i^{\zb^\perp} d\zb = \xb_i^{\yb^\perp}.
\end{align*}
By construction, $\|\xb_i^{\yb^\perp}\|_2> R$ and is non-zero so we conclude that for \eqref{eq:diff_comb} to hold we must have $c_i = 0$. This gives the desired contradiction and implies that $\phi_1, \dots \phi_n$ are independent and $\Vb^\infty$ is positive definite with $\lambda_{\min}(\Vb^{\infty}) = \lambda_0$.\par

Next we consider $\Gb^{\infty}$ and again frame the problem in the context of the covariance matrix of functionals. Define 
\begin{align*}
\theta_i(\vb) \coloneqq \sigma\bigg(\frac{\vb^\top \xb_i}{\|\vb\|_2}\bigg)
\end{align*} for $\vb \ne 0$.

Now the statement of the theorem is equivalent to showing that the covariance matrix of $\{ \theta_i \}$ does not have zero-eigenvalues, that is, the functionals $\theta_i$s are linearly independent. For the sake of contradiction assume $\exists~c_1,\ldots, c_n$ such that \begin{align*}
c_1 \theta_1 + c_2 \theta_2 + \cdots + c_n \theta_n =0\;\;  \text{in}~ \cV \;\;\text{(equivalent to a.e)}.
\end{align*}
Via the same contradiction argument we show that $c_i = 0$ for all $i$.
Unlike $\phi_i$ defined in \eqref{eq:phis}, each $\theta_i$ is continuous and non-negative so equality ``a.e'' is strengthened to ``for all $\vb$'',
\begin{align*}
c_1 \theta_1 + c_2 \theta_2 + \cdots + c_n \theta_n =0. 
\end{align*}
Equality everywhere requires that the derivatives of the function are equal to $0$ almost everywhere. Computing derivatives with respect to $\vb$ yields
\begin{align*}
c_1 \xb_1^{\vb^\perp}\ind \{\vb^\top \xb_1\ge 0 \} + c_2 \xb_2^{\vb^\perp}\ind \{\vb^\top \xb_2\ge 0\} + \cdots + c_n \xb_n^{\vb ^\perp}\ind \{\vb^\top \xb_n \ge 0\} = 0.
\end{align*}
Which coincide with
\begin{align*}
c_1 \phi_1 + \cdots + c_n \phi_n
\end{align*}
By the first part of the proof, the linear combination $c_1 \phi_1 + \cdots + c_n \phi_n$ is non-zero around a ball of positive measure unless $c_i = 0$ for all $i$. This contradicts the assumption that the derivative is $0$ almost everywhere; therefore $\Gb^\infty$ is strictly positive definite with $\lambda_{\min}(G^\infty) \eqqcolon \mu_0 > 0$.
\end{proofLemma}

We briefly derive an inequality for the sum of  indicator functions for events that are bounded by the sum of indicator functions of \emph{independent} events. This enables us to develop more refined concentration than in \citet{du2018gradient} for monitoring the orthogonal and aligned Gram matrices during training. 

\begin{lemma}\label{lemma:bern}
Let $A_{1}, \dots, A_{m}$ be a sequence of events and suppose that $A_{k} \subseteq B_k$ with $B_1, \dots, B_m$ mutually independent. Further assume that for each $k$, $\PP(B_k) \le p$, and define $S = \frac{1}{m} \sum_{k=1}^{m} \ind_{A_k}$. Then with probability $1- \delta$, $S$ satisfies
\begin{align*}
    S \le p \bigg(2 + \frac{8\log(1 / \delta) }{3mp} \bigg). 
\end{align*}
\end{lemma}

\begin{proofLemma}{\ref{lemma:bern}}
Bound $S$ as
\begin{align*}
    S = \frac{1}{m} \sum_{k=1}^{m} \ind_{A_k} \le \frac{1}{m} \sum_{k=1}^{m} \ind_{B_k} . 
\end{align*}
We apply Bernstein's concentration inequality to reach the bound. Denote $X_k = \frac{\ind_{B_k}}{m}$ and $\tilde{S} = \sum_{k=1}^{m} X_{k}$. Then
\begin{align*}
    \var{(X_k)} & \le  \EE X_k^2=  (1/m)^2 \PP(X_k) + 0 \le \frac{p}{m^2}, ~~~ \EE \tilde{S} = \EE \sum_{k=1}^{m}X_k \le p.
\end{align*}
Applying Bernstein's inequality yields
\begin{align*}
\PP(\tilde{S}- \EE \tilde{S} \ge t) & \le  \exp \bigg( \frac{-t^2 /2}{\sum_{k=1}^{m} \EE X_k^2 + \frac{t}{3m}}\bigg).\\
\end{align*}
Fix $\delta$ and take the smallest $t$ such that $\PP(\tilde{S} - \EE\tilde{S} \ge t) \le \delta$. Denote $t = r \cdot \EE \tilde{S}$, either $
\PP(\tilde{S} - \EE \tilde{S}  \ge \EE \tilde{S}) \le \delta$, or $t = r \EE \tilde{S}$  corresponds to $r \ge 1$. Note that $t  = r \EE \tilde{S} \le rp$. In the latter case, the bound is written as
\begin{align*}
\PP(\tilde{S} - \EE \tilde{S} \ge rp) & \le \exp \bigg( \frac{-(pr)^2 /2}{p /m  + \frac{pr}{3m}}\bigg)  
\le \exp \bigg( \frac{-(pr)^2 /2}{ \frac{p}{m}( 1 + \frac{r}{3})}\bigg) 
\le \exp \bigg( \frac{-(pr)^2 /2 }{\frac{p}{m}(\frac{4r}{3}) }\bigg) 
= \exp \bigg( \frac{-3prm}{8}\bigg).
\end{align*}
Solving for $\delta$ gives 
\begin{align*}
rp \le \frac{8\log(1 / \delta) }{3m}. 
\end{align*}
Hence with probability $1-\delta$, 
\begin{align*}
S \le \tilde{S} \le  \max \Bigg\{ p \bigg(1+ \frac{8\log(1/ \delta)}{3mp} \bigg), 2p \Bigg\} \le p \bigg(2+ \frac{8\log(1/ \delta)}{3mp} \bigg). 
\end{align*}
\end{proofLemma}

\begin{proofLemma}{\ref{lemma:hzero}}
We prove the claim by applying concentration on each entry of the difference  matrix. Each entry $\Vb_{ij}(0)$ is written as
\begin{align*}
\Vb_{ij}(0) = \frac{1}{m} \sum_{k=1}^{m}  \big \langle \xb_i^{\vb_k(0)^\perp} ,~ \xb_j^{\vb_k(0)^\perp} \big \rangle \bigg(\frac{\alpha c_k \cdot g_k}{\|\vb_k\|_2}\bigg)^2 \ind_{ik}(0) \ind_{jk}(0).
\end{align*}
At initialization $g_k(0) = \|\vb_k(0)\|_2/ \alpha$,  $c_k^2 = 1$ so $\Vb_{ij}(0)$  simplifies to 
\begin{align*}
\Vb_{ij}(0) = \frac{1}{m} \sum_{k=1}^{m} \big \langle \xb_i^{\vb_k(0)^\perp} ,~ \xb_j^{\vb_k(0)^\perp} \big \rangle \ind_{ik}(0) \ind_{jk}(0).
\end{align*}
Since the weights $\vb_k(0)$ are initialized independently for each entry we have $\EE_{\vb} \Vb_{ij}(0) = \Vb_{ij}^{\infty}$. We measure the deviation $\Vb(0) -\Vb^{\infty}$ via concentration.
Each term in the sum $ \frac{1}{m} \sum_{j=1}^{m} \big \langle \xb_i^{\vb_k(0)^\perp} ,~ \xb_j^{\vb_k(0)^\perp} \big \rangle \ind_{ik}(0) \ind_{jk}(0)$ is independent and bounded,
\begin{align*}-1\le \big \langle \xb_i^{\vb_k(0)^\perp} ,~\xb_j^{\vb_k(0)^\perp} \big \rangle \ind_{ik}(0) \ind_{jk}(0) \le 1.
\end{align*}
Applying Hoeffding's inequality to each entry yields that with probability $1- \delta/n^2$,  for all $i,j$,
\begin{align*}
|\Vb_{ij}(0) -\Vb_{ij}^{\infty}| \le \frac{2 \sqrt{\log(n^2/ \delta) }}{\sqrt{m}}.
\end{align*}
Taking a union bound over all entries, with probability $1 -\delta$, 
\begin{align*}
|\Vb_{ij}(0) -\Vb_{ij}^{\infty}| \le \frac{ 4\sqrt{\log(n/ \delta)}}{\sqrt{m}} .
\end{align*}
Bounding the spectral norm, with probability $1-\delta$,
\begin{align*}
\|\Vb(0) - \Vb^{\infty}\|^2_2 &\le \|\Vb(0) - \Vb^\infty\|_F ^2 \le \sum_{i,j} |\Vb_{ij}(0) -\Vb_{ij}^{\infty}|^2 \\
&\le \frac{16 n^2 \log(n/\delta)}{m}.
\end{align*}

Taking $m  = \Omega \big(\frac{n^2\log(n/ \delta)}{\lambda_0^2} \big)$ therefore guarantees 
\begin{align*}
\|\Vb(0)- \Vb^{\infty}\|_2 \le \frac{\lambda_0}{4}. 
\end{align*}
\end{proofLemma}

\begin{proofLemma}{\ref{lemma:Gzero}}
This is completely analogous to \ref{lemma:hzero}.
Recall $\Gb(0)$ is defined as,
\begin{align*}
\Gb_{ij}(0) = \frac{1}{m} \sum_{k=1}^{m}  \big \langle \xb_i^{\vb_k(0)},~ \xb_j^{\vb_k(0)} \big \rangle c_k^2\ind_{ik}(0) \ind_{jk}(0)
\end{align*}
 with $c_k^2 = 1$ and $\vb_k(0) \sim N(0,\alpha^2 \Ib)$  are initialized i.i.d. Since each term is bounded like \ref{lemma:hzero}. The same analysis gives
\begin{align*}
    \|\Gb_{ij}(0) - \Gb_{ij}^{\infty}\|_2^2 \le \frac{16 n^2 \log(n/\delta)}{m}.
\end{align*}

Taking $m  = \Omega \Big( \frac{n^2\log(n/ \delta)}{\mu_0^2} \Big)$ therefore guarantees,
\begin{align*}
\|\Gb(0)- \Gb^{\infty}\|_2 \le \frac{\mu_0}{4}.
\end{align*}
\end{proofLemma}

\begin{proofLemma}{\ref{lemma:rec}}
For a given $R$, define the event of a possible sign change of neuron $k$ at point $\mathbf{x}_i$ as
\begin{align*}
    A_{i,k}(R) = \{ \exists \vb: \|\vb-\vb_k(0)\|_2 \le R,~\text{and}~ \ind \{ \vb_k(0)^\top \xb_i \ge 0 \}\ne \ind \{ \vb^\top \xb_i \ge 0 \} \}
\end{align*}
$A_{i,k}(R)$ occurs exactly when $| \vb_k (0)^\top \xb_i | \le R$, since $\|\xb_i\|_2=1$ and the perturbation may be taken in the direction of $- \mathbf{x}_i$. To bound the probability $A_{i,k}(R)$ we consider the probability of the event
\begin{align*}
\PP(A_{i,k}(R)) = \PP(|\vb_k(0)^\top \xb_i| < R) = \PP(|z| < R).
\end{align*}
Here, $z~\sim ~ N(0,\alpha^2)$ since the product $\vb_k(0)^{\top}\xb_i$ follows a centered normal distribution. The norm of $\|\xb_i\|_2=1$ which implies that $z$ computes to a standard deviation $\alpha$.
Via estimates on the normal distribution, the probability on the event is bounded like
\begin{align*}
\PP(A_{i,k}(R)) \le \frac{2R}{\alpha \sqrt{2\pi}}.
\end{align*}
We use the estimate for $\PP(A_{i,k}(R))$ to bound the difference between the surrogate Gram matrix and the Gram matrix at initialization $\Vb(0)$.\\
Recall the surrogate $\hat{\Vb}(t)$ is defined as 
\begin{align*}
\hat{\Vb}_{ij}(t) = \frac{1}{m}\sum_{k = 1}^{m} \big \langle \xb_i^{\vb_k(t)^{\perp}},~ \xb_k^{\vb_k(t)^{\perp}}  \big \rangle
\ind_{ik}(t) \ind_{jk}(t).
\end{align*}
Thus for entry $i,j$ we have
\begin{align*}
    |\Hat{\Vb}_{ij}(t) - \Vb_{ij}(0)|&=
    \bigg| \frac{1}{m}   \sum_{k=1}^{m}\big \langle \xb_i^{\vb_k(t)^\perp},~ \xb_j^{\vb_k(t)^\perp} \big \rangle \ind_{ik}(t)\ind_{jk}(t)- \langle \xb_i^{\vb_k(0)^\perp},~ \xb_j^{\vb_k(0)^\perp} \rangle  \ind_{ik}(0)\ind_{jk}(0) \bigg|
\end{align*}

This sum is decomposed into the difference between the inner product and the difference in the rectifier patterns terms respectively: \begin{align*}
    &\bigg(\big\langle\xb_i^{\vb_k(t)^\perp}, \xb_j^{\vb_k(t)^\perp} \big\rangle- \big\langle \xb_i^{\vb_k(0)^\perp}, \xb_j^{\vb_k(0)^\perp} \big\rangle \bigg),\qquad 
\bigg( \ind_{ik}(t)\ind_{jk}(t) - \ind_{ik}(0)\ind_{jk}(0)  \bigg).\end{align*} 
Define
\begin{align*}
    Y_{ij}^k&=\bigg(\big\langle\xb_i^{\vb_k(t)^\perp},~\xb_j^{\vb_k(t)^\perp} \big\rangle- \big\langle \xb_i^{\vb_k(0)^\perp},~\xb_j^{\vb_k(0)^\perp} \big\rangle \bigg)\big(\ind_{ik}(t) \ind_{jk}(t) \big),\\
Z_{ij}^k &= \bigg(\big\langle \xb_i^{\vb_k(0)^\perp},~\xb_j^{\vb_k(0)^\perp} \big\rangle \bigg)  \bigg( \ind_{ik}(t)\ind_{jk}(t) - \ind_{ik}(0)\ind_{jk}(0) \bigg).
\end{align*}
Then 
\begin{align*}
|\Hat{\Vb}_{ij}(t) - \Vb_{ij}(0)|  = \bigg| \frac{1}{m} \sum_{k=1}^{m} Y_{ij}^k+ Z_{ij}^k \bigg| \le \bigg| \frac{1}{m} \sum_{k=1}^{m} Y_{ij}^k \bigg| + \bigg| \frac{1}{m} \sum_{k=1}^{m} Z_{ij}^k \bigg|. 
\end{align*}
To bound $|\frac{1}{m} \sum_{k=1}^{m} Y_{ij}^k |$ we bound each $|Y^k_{ij}|$ as follows.

\begin{align*}
    |Y^k_{ij}| &= \Bigg| \bigg(\big\langle\xb_i^{\vb_k(t)^\perp},~\xb_j^{\vb_k(t)^\perp} \big\rangle- \big\langle \xb_i^{\vb_k(0)^\perp},~\xb_j^{\vb_k(0)^\perp} \big\rangle \bigg)\big(\ind_{ik}(t) \ind_{jk}(t) \big) \Bigg| \\
    &\le  \bigg| \big\langle\xb_i^{\vb_k(t)^\perp},~\xb_j^{\vb_k(t)^\perp} \big\rangle- \big\langle \xb_i^{\vb_k(0)^\perp},~\xb_j^{\vb_k(0)^\perp} \big\rangle \bigg| \\
  &= \bigg|\langle \xb_i, \xb_j \rangle -  \big\langle\xb_i^{\vb_k(t)},~\xb_j^{\vb_k(t)} \big\rangle + \big\langle \xb_i^{\vb_k(0)},~\xb_j^{\vb_k(0)} \big\rangle - \langle \xb_i, \xb_j \rangle \bigg| \\
  &= \Bigg|  \bigg\langle \frac{\xb_i^\top \vb_k(t)}{\|\vb_k(t)\|_2}\cdot \frac{ \vb_k(t)}{\|\vb_k(t)\|_2},~\frac{\xb_j^\top \vb_k(t)}{\|\vb_k(t)\|_2}\cdot \frac{ \vb_k(t)}{\|\vb_k(t)\|_2} \bigg\rangle - \big\langle \xb_i^{\vb_k(0)},~\xb_j^{\vb_k(0)} \big\rangle \Bigg| \\
  &= \Bigg| \frac{\xb_i^\top \vb_k(t)}{\|\vb_k(t)\|_2}\cdot \frac{\xb_j^\top \vb_k(t)}{\|\vb_k(t)\|_2} - \big\langle \xb_i^{\vb_k(0)},~\xb_j^{\vb_k(0)} \big\rangle \Bigg| \\
  &= \Bigg|  \frac{\xb_i^\top \vb_k(0)}{\|\vb_k(0)\|_2}\cdot \frac{\xb_j^\top \vb_k(0)}{\|\vb_k(0)\|_2}+ \xb_i^\top\bigg(\frac{\vb_k(t)}{\|\vb_k(t)\|_2} - \frac{\vb_k(0)}{\|\vb_k(0)\|_2}\bigg)\cdot \frac{\xb_j^\top \vb_k(t)}{\|\vb_k(t)\|_2}\\
  &\qquad+\xb_j^\top\bigg(\frac{\vb_k(t)}{\|\vb_k(t)\|_2} - \frac{\vb_k(0)}{\|\vb_k(0)\|_2}\bigg)\cdot \frac{\xb_i^\top \vb_k(0)}{\|\vb_k(0)\|_2}
   - \big\langle \xb_i^{\vb_k(0)},~\xb_j^{\vb_k(0)} \big\rangle \Bigg|\\
  &\le \Bigg| \xb_i^\top\bigg(\frac{\vb_k(t)}{\|\vb_k(t)\|_2} - \frac{\vb_k(0)}{\|\vb_k(0)\|_2}\bigg)\cdot \frac{\xb_j^\top \vb_k(t)}{\|\vb_k(t)\|_2}\Bigg|+  \Bigg| \xb_i^\top\bigg(\frac{\vb_k(t)}{\|\vb_k(t)\|_2} - \frac{\vb_k(0)}{\|\vb_k(0)\|_2}\bigg)\cdot \frac{\xb_j^\top \vb_k(t)}{\|\vb_k(t)\|_2}\Bigg|\\
  &\le 2\bigg \|\frac{\vb_k(t)}{\|\vb_k(t)\|_2} - \frac{\vb_k(0)}{\|\vb_k(0)\|_2} \bigg\|_2.
\end{align*}
Therefore, we have
\begin{align*}
\bigg|\frac{1}{m} \sum_{k=1}^{m} Y_{ij}^k \bigg|
&\leq \frac{2}{m}\sum_{k=1}^{m} \bigg \|\frac{\vb_k(t)}{\|\vb_k(t)\|_2} - \frac{\vb_k(0)}{\|\vb_k(0)\|_2} \bigg\|_2 \\
&\le \frac{4R_{\vb}(2m /\delta)^{1/d}}{\alpha}\\
&\le \frac{8R_{\vb}(m /\delta)^{1/d}}{\alpha},
\end{align*}
where the first inequality follows from Lemma \ref{lemma:mdelta}. Note that the inequality holds with high probability $1- \delta /2$ for all $i,j$. 

For the second sum, $|\frac{1}{m} \sum_{k=1}^{m} Z_{ij}^k| \le \frac{1}{m}\sum_{k=1}^{m} \ind_{A_{ik}(R)} +  \frac{1}{m}\sum_{k=1}^{m} \ind_{A_{jk}(R)}$ so we apply Lemma \ref{lemma:bern} to get, with probability $1-\delta/ 2n^2$
 \begin{align*}
     \bigg|\frac{1}{m} \sum_{k=1}^{m} Z_{ij}^k \bigg| &\le  \frac{2R_{v}}{\alpha \sqrt{2 \pi }}\bigg( 2 + \frac{2\sqrt{2\pi}\alpha \log{(2n^2/\delta)}}{3mR_v}\bigg) \\
     &\le \frac{8R_{v}}{\alpha \sqrt{2\pi}} , 
 \end{align*}
since $m$ satisfies $m = \Omega \big( \frac{(m /\delta)^{1 /d}n^2 \log(n / \delta) }{\alpha \lambda_0}\big)$.
Combining the two sums for $Y_{ij}^k$ and $Z_{ij}^k$, with probability $1 -\frac{\delta}{2n^2}$, 
\begin{align*}
|\hat{\Vb}_{ij}(t) - \Vb_{ij}(0)| \le \frac{8R_v}{\alpha \sqrt{2 \pi}} + \frac{8R_{v}(m /\delta)^{1/d}}{\alpha} \le \frac{12R_v(m/\delta)^{1/d}}{\alpha}.
\end{align*}
Taking a union bound, with probability $1-\delta/2$,
\begin{align*}
\|\hat{\Vb}(t) - \Vb(0)\|_{F}
 = \sqrt{\sum_{i,j} |\Hat{\Vb}_{ij}(t) - \Vb_{ij}(0)|^2} \le   \frac{12nR_v(m/\delta)^{1/d}}{\alpha}.
\end{align*}
Bounding the spectral norm by the Frobenous norm,
\begin{align*}
\|\hat{\Vb}(t) - \Vb(0) \|_2 \le \frac{12nR_v(m/\delta)^{1/d}}{\alpha}.
\end{align*}
Taking $R_v = \frac{ \alpha \lambda_0 }{96n(m / \delta)^{1 /d} }$ gives the desired bound.
\begin{align*}
\|\hat{\Vb}(t) - \Vb(0) \|_2 \le \frac{\lambda_0}{8}.
\end{align*}
\end{proofLemma}

\begin{proofLemma}{\ref{lemma:ht}}
To bound $\|\Vb(t) - \Vb(0)\|_2$
we now consider $\|\Vb(t) - \hat{\Vb}(t)\|_2$. The entries of 
$\Vb_{ij}(t)$ are given as
\begin{align*}\Vb_{ij}(t) = \frac{1}{m} \sum_{k = 1}^{m}  \big\langle \xb_i^{\vb_k(t)^{\perp}},~ x_{j}^{\vb_k(t)^{\perp}} \big \rangle
\ind_{ik}(t) \ind_{jk}(t) \bigg(\frac{\alpha c_k\cdot g_k}{\|\vb_k(0)\|_2}\bigg)^{2}. \end{align*}
The surrogate $\hat{\Vb}(t)$  is defined as
\begin{align*}
   \hat{\Vb}_{ij}(t)&= \frac{1}{m} \sum_{k = 1}^{m}  \big \langle \xb_i^{\vb_k(t)^{\perp}},~ x_{j}^{\vb_k(t)^{\perp}} \big \rangle
\ind_{ik}(t) \ind_{jk}(t).
\end{align*}

The only difference is in the second layer terms. The difference between each entry is written as
\begin{align*}
|\Vb_{ij}(t) - \hat{\Vb}_{ij}(t) |  &= \bigg|\frac{1}{m} \sum_{k = 1}^{m} \big \langle \xb_i^{\vb_k(t)^{\perp}},~ x_{j}^{\vb_k(t)^{\perp}} \big \rangle
\ind_{ik}(t) \ind_{jk}(t)\Bigg(\bigg(\frac{ \alpha c_k\cdot g_k}{\|\vb_k(t)\|_2}\bigg)^{2} -1\Bigg)\Bigg | \\
&\le \max_{1 \le k \le m} \bigg(\frac{\alpha^2  g_k(t)^2}{\|\vb_k(t)\|_2^2} -1\bigg).
\end{align*}
Write $1 = \frac{\alpha^2 g_k^2(0)}{\|\vb_k(0)\|^2_2}$, since $\|\vb_k(t)\|_2$ is increasing in $t$ according to (\ref{eq:inc})
\begin{align*}
\frac{\alpha^2  g_k(t)^2}{\|\vb_k(t)\|_2^2} -1 =  \frac{\alpha^2  g_k(t)^2}{\|\vb_k(t)\|_2^2} - \frac{\alpha^2 g_k(0)^2}{\|\vb_k(0)\|^2_2} \le 3R_g(m/\delta)^{1 /d}  + 3R_{v}(m/\delta)^{1/d}/\alpha.
\end{align*}
The above inequality is shown by considering different cases for the sign of the difference $g_k(t) -g_k(0)$. 
Now 
\begin{align*}
   \Bigg| \frac{\alpha^2  g_k(t)^2}{\|\vb_k(t)\|_2^2} - \frac{\alpha^2 g_k(0)^2}{\|\vb_k(0)\|^2_2} \Bigg| &= \Bigg| \bigg( \frac{\alpha g_k(t)}{\|\vb_k(t)\|_2} + \frac{\alpha g_k(0) }{\|\vb_k(0)\|_2}\bigg)\bigg( \frac{\alpha g_k(t)}{\|\vb_k(t)\|_2} - \frac{\alpha g_k(0) }{\|\vb_k(0)\|_2}\bigg) \Bigg|  \\
   &\le \Bigg| \bigg( \frac{\alpha g_k(0) + \alpha R_g}{\|\vb_k(0)\|_2} + \frac{\alpha g_k(0) }{\|\vb_k(0)\|_2}\bigg)\bigg( \frac{\alpha g_k(t)}{\|\vb_k(t)\|_2} - \frac{\alpha g_k(0) }{\|\vb_k(0)\|_2}\bigg) \Bigg| \\
   &\le (2+ R_g(m /\delta)^{1/d})\Bigg| \bigg( \frac{\alpha g_k(t)}{\|\vb_k(t)\|_2} - \frac{\alpha g_k(0) }{\|\vb_k(0)\|_2}\bigg) \Bigg| \\
   &\le (2+ R_g(m /\delta)^{1/d}) \max\Bigg( \bigg|\frac{\alpha(g_k(0) + R_g)}{\|\vb_k(0)\|_2} - \frac{\alpha g_k(0)}{\|\vb_k(0)\|_2} \bigg|, \bigg| \frac{\alpha(g_k(0) -R_g)}{\|\vb_k(0)\|_2 + R_v} - \frac{\alpha g_k(0)}{\|\vb_k(0)\|_2}\bigg| \Bigg) \\
   &\le (2+ R_g(m /\delta)^{1/d})\max\big(R_g(m/\delta)^{1/d}, R_g(m/\delta)^{1/d}+  R_v(m/\delta)^{1/d}/\alpha \big)\\
   &\le 3R_g(m/\delta)^{1/d}+ 3R_v(m/\delta)^{1/d}/ \alpha,
\end{align*}
where the second inequality holds due to Lemma \ref{lemma:mdelta} with probability $1- \delta$  over the initialization.

Hence:
\begin{align*}
\|\hat{\Vb}(t) - \Vb(t) \|_2  \le \| \hat{\Vb}(t) - \Vb(t) \|_{F} = \sqrt
{\sum_{i,j} |\hat{\Vb}_{ij}(t) - \Vb_{ij}(t) |^2 }\le 3n R_g (m / \delta)^{1 /d} +  3nR_{v}(m/\delta)^{1/d}/\alpha.
\end{align*}
Substituting $R_v, R_g$ gives
\begin{align*}
\|\hat{\Vb}(t) - \Vb(t) \|_2 \le \frac{\lambda_0}{8}.
\end{align*}

Now we use Lemma \ref{lemma:rec} to get that with probability $1-\delta$
\begin{align*}
\|\hat{\Vb}(t) - \Vb(0) \|_2 \le \frac{\lambda_0}{8}. 
\end{align*}
Combining, we get with probability $1-\delta$
\begin{align*}
\|\Vb(t)  -\Vb(0) \|_2\le \frac{\lambda_0}{4}.
\end{align*}
We note that the source for all the high probability uncertainty $1-\delta$ all arise from initialization and the application of Lemma \ref{lemma:mdelta}.
\end{proofLemma}

\begin{proofLemma}{\ref{lemma:Gt}}
To prove the claim we consider each entry $i,j$ of $\Gb(t)- \Gb(0)$. We have,
\begin{align*}
|\Gb_{ij}(t) - \Gb_{ij}(0)|&= \Bigg|\frac{1}{m}   \sum_{k=1}^{m}
\sigma \bigg(\frac{\vb_k(t)^\top \xb_i}{\|\vb_k(t)\|_2}\bigg)\sigma \bigg(\frac{\vb_k(t)^\top \xb_j}{\|\vb_k(t)\|_2}\bigg) -
\sigma \bigg(\frac{\vb_k(0)^\top \xb_i}{\|\vb_k(0)\|_2}\bigg)\sigma \bigg(\frac{\vb_k(0)^\top \xb_j}{\|\vb_k(0)\|_2}\bigg) \Bigg| \\
 & \le
\frac{1}{m}  \Bigg| \sum_{k=1}^{m}
\sigma \bigg(\frac{\vb_k(t)^\top \xb_i}{\|\vb_k(t)\|_2}\bigg)\sigma \bigg(\frac{\vb_k(t)^\top \xb_j}{\|\vb_k(t)\|_2}\bigg) -
\sigma \bigg(\frac{\vb_k(t)^\top \xb_i}{\|\vb_k(t)\|_2}\bigg)\sigma \bigg(\frac{\vb_k(0)^\top \xb_j}{\|\vb_k(0)\|_2}\bigg) \Bigg| \\
& \qquad+
\frac{1}{m}  \Bigg| \sum_{k=1}^{m}
\sigma \bigg(\frac{\vb_k(t)^\top \xb_i}{\|\vb_k(t)\|_2}\bigg)\sigma \bigg(\frac{\vb_k(0)^\top \xb_j}{\|\vb_k(0)\|_2}\bigg) -
\sigma \bigg(\frac{\vb_k(0)^\top \xb_i}{\|\vb_k(0)\|_2}\bigg)\sigma \bigg(\frac{\vb_k(0)^\top \xb_j}{\|\vb_k(0)\|_2}\bigg) \Bigg| \\
 & \le 2\bigg\| \frac{\vb_k(t)}{\|\vb_k(t)\|_2} - \frac{\vb_k(0)}{\|\vb_k(0)\|_2} \bigg\|_2 \le \frac{2\tilde{R}_v (m / \delta)^{1 /d} }{\alpha}.
\end{align*}
In the last inequality we used the fact that 
\begin{align*}
\bigg\|\frac{\vb_k(0)}{\|\vb_k(0)\|_2} - \frac{\vb_k(t)}{\|\vb_k(t)\|_2}\bigg\|_2 \le \frac{ \|\vb_k(t) - \vb_k(0)\|_2 }{\|\vb_k(0)\|_2}\le \frac{(m /\delta)^{1/ d}}{\alpha} \|\vb_k(t) - \vb_k(0)\|_2,
\end{align*}
where the first inequality uses that $\|\vb_k(0)\|_2 \le \|\vb_k(t)\|_2$ 
and is intuitive from a geometrical standpoint. 
Algebraically given vectors $\ab, \bbb$, then for any $c\ge 1$
\begin{align*}
    \bigg\|\frac{\ab c}{\|\ab\|_2} -  \frac{\bbb}{\|\bbb\|_2}\bigg\|^2_2 &=   \bigg\|\frac{\ab}{\|\ab\|_2}  - \frac{\bbb}{\|\bbb\|_2} + (c-1)\frac{\ab}{\|\ab\|_2}\bigg\|_2^2 \\
    &= \bigg\|\frac{\ab}{\|\ab\|_2} -  \frac{\bbb}{\|\bbb\|_2}\bigg\|^2_2 + (c-1)^2 + 2(c-1) \bigg\langle \frac{\ab}{\|\ab\|_2}  - \frac{\bbb}{\|\bbb\|_2}, \frac{\ab}{\|\ab\|_2}  \bigg\rangle \\
    &\ge \bigg\|\frac{\ab}{\|\ab\|_2} -  \frac{\bbb}{\|\bbb\|_2}\bigg\|^2_2 + (c-1)^2 \ge \bigg\|\frac{\ab}{\|\ab\|_2} -  \frac{\bbb}{\|\bbb\|_2}\bigg\|^2_2.
\end{align*}
The first inequality in the line above is since $\frac{\langle \ab,\bbb\rangle }{\|\ab\|_2, \|\bbb\|_2}\le 1.$\par
Hence,
\begin{align*}
\|\Gb(t) - \Gb(0) \|_2\le \|\Gb(t) - \Gb(0)\|_F =  \sqrt{\sum_{i,j} | \Gb_{ij}(t) - \Gb_{ij}(0)|^2} \le  \frac{2n\tilde{R}_v  (m / \delta)^{1 /d}}{\alpha \sqrt{2\pi}}.
\end{align*}

Taking $\tilde{R}_v = \frac{\sqrt{2\pi}\alpha  \mu_0}{8n (m /\delta)^{1/d}}$ gives the desired bound. Therefore, with probability $1-\delta$,
\begin{align*}
    \|\Gb(t) - \Gb(0) \|_2 \le \frac{\mu_0}{4}.
\end{align*}
\end{proofLemma}

Now that we have established bounds on $\Vb(t), \Gb(t)$ given that the parameters stay near initialization, we show that the evolution converges in that case:

\begin{proofLemma}{\ref{lemma:exp}}
Consider the squared norm of the predictions $\|\fb(t) -\yb\|_2^2$.
Taking the derivative of the loss with respect to time,
\begin{align*}
\frac{d}{dt}\| \fb (t)-\yb \|^2_2 = -2(\fb (t) - \yb)^{\top} \bigg(\Gb(t) + \frac{\Vb(t)}{\alpha^2} \bigg)(\fb (t) - \yb).
\end{align*}
Since we assume that $\lambda_{\min}\bigg( \Gb(t) + \frac{\Vb(t)}{\alpha^2} \bigg) \ge \frac{\omega}{2}$, the derivative of the squared norm is bounded as 
\begin{align*}
\frac{d}{dt}\|\fb (t) - \yb\|_2^2 \le -\omega \|\fb (t)- \yb\|_2^2.
\end{align*}
Applying an integrating factor yields
\begin{align*}
\|\fb (t) - \yb\|_2^2 \exp(\omega t) \le C. 
\end{align*}
Substituting the initial conditions, we get
\begin{align*}
\|\fb (t) - \yb\|_2^2 \le \exp(-\omega t) \|\fb (0) -\yb \|_2^2.
\end{align*}
\end{proofLemma}

For now, assuming the linear convergence derived in Lemma \ref{lemma:exp},  we bound the distance of the parameters from initialization. Later we combine the bound on the parameters and Lemmas \ref{lemma:Vt}, \ref{lemma:Gt} bounding the least eigenvalue of $\bLambda(t)$, to derive a condition on the over-parametrization $m$ and ensure convergence from random initialization.

\begin{proofLemma}{\ref{lemma:closew}} 
Denote $f(\xb_i)$ at time $t$ as $f_i(t)$.
Since $\|\xb_i^{\vb_k(t)^\perp}\|_2\le \|\xb_i\|_2= 1$, we have that
\begin{align*}
\bigg\|\frac{d \vb_k(t)}{dt} \bigg\|_2 &=\bigg\| \sum_{i =1 }^{n}(y_i -f_i(t))\frac{1}{\sqrt{m}}c_{k}g_k(t) \frac{1}{\|\vb_k(t)\|_2} \xb_i^{\vb^\perp} \ind_{ik}(t)\bigg\|_2\\
& \le \frac{1}{\sqrt{m}}\sum_{i=1}^{n} |y_i - f_i(t)| \frac{c_k g_k(t)}{\|\vb_k(t)\|_2} .
\end{align*}
Now using \eqref{eq:inc} and the initialization  $\|\vb_k(0)\|= \alpha g_k(0)$, we bound $\bigg|\frac{c_k g_k(t)}{\|\vb_k(t)\|_2} \bigg|$, 
\begin{align*}
\bigg|\frac{c_k g_k(t)}{\|\vb_k(t)\|_2} \bigg| \le \bigg|c_k  \bigg( \frac{g_k(0) + R_g}{\|\vb_k(0)\|_2} \bigg) \bigg| \le \frac{1}{\alpha} \bigg(1 +  \alpha R_g / \|\vb_k(0)\|_2\bigg).
\end{align*}
By Lemma \ref{lemma:mdelta}, we have that with probability $1-\delta$ over the initialization, 
\begin{align*} 
\alpha / \|\vb_k(0)\|_2\le C(m / \delta)^{1 / d} .
\end{align*}
Hence $\alpha R_g / \|\vb_k(0)\|_2\le 1$. This fact bounds $\bigg|\frac{c_k g_k(t)}{\|\vb_k(t)\|_2} \bigg| $ with probability $1-\delta$ for each $k$,
\begin{align*}
\bigg| \frac{c_k g_k(t)}{\|\vb_k(t)\|_2} \bigg| \le 2 /\alpha.
\end{align*}
Substituting the bound,
\begin{align*}
\bigg \| \frac{d}{dt}\vb_k(t) \bigg \|_2&\le \frac{2}{\alpha \sqrt{m}} \sum_{i=1}^{n}|f_i(t) - y_i| \\
&\le \frac{2\sqrt{n}}{\alpha \sqrt{m}}\|\fb (t) - \yb\|_2 \\
&\le \frac{2\sqrt{n}}{\alpha \sqrt{m}}\exp(-\omega t /2)\|\fb (0) - \yb\|_2.
\end{align*}
Thus, integrating and applying Jensen's inequality, 
\begin{align*}
\|\vb_k(t) -\vb_k(0)\|_2 \le \int_0^s \bigg \|\frac{d \vb_k(s)}{dt} \bigg\|_2 ds \le \frac{4\sqrt{n}\|\fb (0) - \yb\|_2}{\alpha \omega \sqrt{m}}.
\end{align*}
Note that the condition $|g_k(t)-g_k(0)| \le R_g$ is stronger than needed and merely assuring that $|g_k(t) - g_k(0)| \le 1/(m /\delta)^{1 / d}$ suffices.
\end{proofLemma} 

Analogously we derive bounds for the distance of $g_k$ from initialization.

\begin{proofLemma}{\ref{lemma:closeg}}
Consider the magnitude of the derivative $\frac{dg_k}{dt}$,
\begin{align*}
\bigg| \frac{dg_k}{dt} \bigg| = \bigg| \frac{1}{\sqrt{m}} \sum_{j = 1}^{n} (f_j - y_j)\frac{c_k}{\|\vb_k\|_2} \sigma(\vb_k^{\top}\xb_j) \bigg|.
\end{align*} Note
\begin{align*}
\bigg|\frac{c_k}{\|\vb_k\|_2}\sigma(\vb_k^\top \xb_j) \bigg|= \bigg|\sigma \bigg(\frac{\vb_k^\top \xb_j}{\|\vb_k\|_2} \bigg) \bigg|
 \le 1
\end{align*}
Thus applying Cauchy Schwartz
\begin{align*}
\bigg| \frac{dg_k(t)}{dt} \bigg| \le \frac{2\sqrt{n}}{\sqrt{m}} \|\fb (t) - \yb\|_2 \le \frac{2\sqrt{n}}{\sqrt{m}} \exp(-\omega t /2 ) \|\fb (0) - \yb \|_2,
\end{align*}
and integrating from $0$ to $t$ yields
\begin{align*}
|g_k(t) - g_k(0)| \le \int_{0}^{t} \bigg| \frac{dg_k}{dt}(s) \bigg| ds \le \int_0 ^t \frac{2\sqrt{n}}{\sqrt{m}} \exp(-\omega s /2 ) \|\fb (0) - \yb \|_2 ds \le \frac{4\sqrt{n}\|\yb - \fb (0)\|_2}{\sqrt{m} \omega}.
\end{align*}
\end{proofLemma}

\begin{proofLemma}{\ref{lemma:init}}
Consider the $i$th entry of the network at initialization, 
\begin{align*}
f_i(0) = \frac{1}{\sqrt{m}} \sum_{k=1}^{m} c_k \sigma \bigg( \frac{g_k \vb_k^\top \xb_i}{\|\vb_k\|_2}\bigg).
\end{align*}
Since the network is initialized randomly and $m$ is taken to be large we apply concentration to bound $f_i(0)$ for each $i$.
Define $z_k =  c_k \sigma \bigg( \frac{g_k(0) \vb_k(0)^\top \xb_i}{\|\vb_k(0)\|_2}\bigg)$. Note that $z_k$ are independent sub-Gaussian random variables with 
\begin{align*}
\|\zb_k\|_{\psi} \le \|N(0,1) \|_{\psi} = C.
\end{align*}
Here $\|\cdot\|_{\psi}$ denotes the 2-sub-Gaussian norm, (see \cite{vershynin2018high} for example). Applying Hoeffding's inequality bounds $f_i(0)$ as
\begin{align*}
\PP( |\sqrt{m}f_i(0)| > t) &\le 2\exp\bigg( -\frac{t^2/2}{\sum_{k=1}^{m} \|\zb_k\|_{\psi_2}}\bigg)  \\
& = 2 \exp \bigg( \frac{-t^2}{2mC} \bigg) .
\end{align*}
Which gives with probability $1 - \delta / n$ that 
\begin{align*}
    |f_i(0) | \le \tilde{C}\sqrt{\log{(n /\delta)}}.
\end{align*}
Now with probability $1- \delta$ we have that, for each $i$, 
\begin{align*}
    |f_i(0) - y_i|  \le |y_i| + \tilde{C} \sqrt{\log(n / \delta)} \le C_2  \sqrt{\log(n / \delta)}.
\end{align*}
Since $y_i = O(1)$. Hence, with probability $1- \delta$,
\begin{align*}
    \|\fb(0) - \yb \|_2 \le C\sqrt{n \log( n /\delta)}.
\end{align*}
\end{proofLemma} 

\begin{proofLemma}{\ref{lemma:mdelta}}
At initialization $\vb_k \sim N(0,\alpha^2 \Ib)$ so the norm behaves like $\|\vb_k(0)\|^2_2\sim \alpha^2 \chi_d$. The cumulative density of a chi-squared distribution with $d$ degrees of freedom behaves like $F(x) = \Theta(x^{d/2})$ for small $x$ so we have that with probability $1 - \frac{\delta}{m}$, that $\|\vb_k(0)\|_2 \ge \alpha (m /\delta)^{\frac{1}{d}}$ where $d$ is the input dimension. Applying a union bound, with probability $1- \delta$, for all $1 \le k \le m$, 
\begin{align*}
\frac{1}{\|\vb_k(0)\|_2} \le \frac{ \big( m / \delta \big)}{\alpha}^{1/d}.
\end{align*}
Now by \eqref{eq:inc} for $t \ge 0$, $\|\vb_k(t) \|_2 \ge \|\vb_k(0)\|_2$ so
\begin{align*}
\frac{1}{\|\vb_k(t)\|_2} \le \frac{1}{\|\vb_k(0)\|_2}  \le \frac{ \big( m / \delta \big)}{\alpha}^{1/d}.
\end{align*}
\end{proofLemma}

\section{Proofs of Lemmas from Appendix \ref{sec:finite_step} and Proposition \ref{theorem:fast_WN}} \label{appendix:finite_step_lemmas}

\begin{proofProp}{\ref{theorem:fast_WN}}
The proof of proposition 2, follows the proofs of Theorems \ref{theorem:finite_step_V}, \ref{theorem:finite_step_G}, and relies on Theorem \ref{theorem:finite_step_proof}. 
In particular for each $\alpha>0$ at initialization, take $\omega_{\alpha}(s) = \lambda_{\min}(\bLambda(s))$ and define the auxiliary $\omega_{\alpha,0} = \lambda_{\min}(\Vb^{\infty} /\alpha^2 + \Gb^{\infty})$. Then we have that 
\begin{align*}
\omega_{\alpha,0} \ge \lambda_0 /\alpha^2 + \mu_0  > 0.    
\end{align*}
Hence, by the same arguments of Theorem 4.1, 4.2 for $\omega_{\alpha}(s)$ if $m = \big(n^4 \log(n/ \delta) / \alpha^4 \omega_{\alpha,0}^4 \big)$, then we have that the conditions of Theorem \ref{theorem:finite_step_proof} are satisfied, namely, $\lambda(s) \ge \frac{\lambda_0}{2}$ and $\mu(s) \ge \frac{\mu_0}{2}$. Taking $\eta_{\alpha} = O\bigg(\frac{1}{\|\bLambda(s)\|_2}\bigg)$, then the required step-size for convergence is satisfied. This follows from the same argument of Theorems \ref{theorem:finite_step_V}, \ref{theorem:finite_step_G} and depends on the fact that $\|\bLambda(s) -\bLambda(0) \|_2 \le \frac{1}{\alpha^2} \|\Vb(s) - \Vb^{\infty}(0)\|_2 + \|\Gb(s) - \Gb(0) \|_2$. 
Now we consider the term, $\alpha \omega_{\alpha, 0}$. For $\alpha=1$,
\begin{align*}
    \alpha \omega_{\alpha,0} = \lambda_{\min}(\Hb^{\infty}).
\end{align*}
Which matches the results of un-normalized convergence. In general, we have that
\begin{align*}
     \alpha \omega_{\alpha,0} \ge \alpha (\lambda_0 /\alpha^2 + \mu_0) \ge \min\{\lambda_0, \mu_0\}.
\end{align*}
Therefore the bound on $m$ is taken to be independent of $\alpha$ as
$m = \Omega \bigg( \frac{n^4 \log(n/\delta) }{\min\{\mu_0^4, \lambda_0^4 \}} \bigg)$
which simplifies the presentation. Now for each $\alpha$ the effective convergence rate is dictated by the least eigenvalue $\omega_{\alpha}$ and the allowed step-size $\eta_{\alpha}$ as,
\begin{align*}
    \bigg( 1 - \eta_{\alpha} \omega_{\alpha} \bigg). 
\end{align*}
Then taking $\alpha^{*} = \argmin_{\alpha > 0} ( 1 - \eta_{\alpha} \omega_{\alpha})$
we have that 
\begin{align*}
    ( 1 - \eta_{\alpha^{*}} \omega_{\alpha^{*}} ) \le ( 1 - \eta_{1} \omega_{1}).
\end{align*}
which corresponds to the un-normalized converegence rate. Therefore as compared with un-normalized training we have that for $\alpha^*$, WN enables a faster convergence rate.
\end{proofProp}

\begin{proofLemma}{\ref{lemma:seminar}}
 Fix $R$, without the loss of generality we write $S_i$ for $S_i(R)$.
 For each $k$, $\vb_k(0)$ is initialized independently via $\sim N(0, \alpha^2\Ib)$, and for a given $k$, the event $\ind_{ik}(0) \ne \ind\{\vb^\top \xb_i \ge 0\}$ corresponds to $|\vb_k(0)^\top \xb_i| \le R$. Since $\|\xb_i\|_2=1$,
 $\vb_k(0)^\top \xb_i ~\sim N(0,\alpha^2)$. Denoting the event that an index $k \in S_i$ as $A_{i,k}$, we have
 \begin{align*}
 \PP(A_{i,k}) \le \frac{2R}{\alpha \sqrt{2\pi}}.
 \end{align*}
 Next the cardinality of $S_i$ is written as
 \begin{align*}
     |S_i| = \sum_{k=1}^{m} \ind_{A_{i,k}}. 
 \end{align*} 
Applying Lemma \ref{lemma:bern}, with probability  $1- \delta /n$, 
\begin{align*}
    |S_i| \le \frac{2mR}{\alpha \sqrt{2 \pi}} +  \frac{16\log(n / \delta) }{3}.
\end{align*}
Taking a union bound, with probability $1-\delta$, for all $i$ we have that
\begin{align*}
    |S_i| \le \frac{2mR}{\alpha \sqrt{2 \pi}} +  \frac{16\log(n / \delta) }{3}.
\end{align*}
 \end{proofLemma}

\begin{proofLemma}{\ref{lemma:gclose_omega}}
To show this we bound the difference $g_k(s) -g_k(0)$ as the sum of the iteration updates. Each update is written as 
\begin{align*}
\bigg|\pdv{L(s)}{g_k}\bigg| = \bigg| \frac{1}{\sqrt{m}}\sum_{i =1 }^{n}(f_i (s) - y_i)
 \frac{c_k}{\|\vb_k(s)\|_2}\sigma(\vb_k(s)^\top \xb_i) \bigg|.
\end{align*} 
As $ \bigg|c_k \sigma \bigg(\frac{\vb_k(s)^\top \xb_i}{\|\vb_k(s)\|_2}\bigg) \bigg| \le 1$,
\begin{align*}
\bigg|\pdv{L(s)}{g_k}\bigg| \le \frac{1}{\sqrt{m}}\sum_{i}^{n}|f_i(s) - y_i|  \le \frac{\sqrt{n}}{\sqrt{m}}\|\fb (s) - \yb\|_2 . 
\end{align*}
By the assumption in the statement of the lemma, 
\begin{align*}
\bigg|\pdv{L(s)}{g_k}\bigg| \le \frac{\sqrt{n} (1 - \frac{\eta \omega}{2})^{s/2}\|\fb (0) - \yb\|_2}{\sqrt{m}}.
\end{align*}
Hence bounding the difference by the sum of the gradient updates:
\begin{align*}
|g_k(K+1) - g_k(0)| \le \eta \sum_{s=0}^{K} \bigg|\pdv{L(s)}{g_k} \bigg|  \le  \frac{4\eta \sqrt{n}\|\fb (0) - \yb\|_2  }{\sqrt{m}}\sum_{s=0}^{K}(1 - \frac{\eta \omega}{2})^{s/2}.
\end{align*}
The last term yields a geometric series that is bounded as
\begin{align*}
\frac{1}{1 -\sqrt{1 - \frac{\eta \omega}{2}}} \le \frac{4}{\eta \omega},
\end{align*}
Hence
\begin{align*}
    |g_k(K+1) - g_k(0)| \le  \frac{4\sqrt{n}\|\fb (0) - \yb\|_2  }{\omega \sqrt{m} }.\end{align*}
\end{proofLemma} 

\begin{proofLemma}{\ref{lemma:finite_step_closev}}

To show this we write $\vb_k(s)$ as the sum of gradient updates and the initial weight $\vb_k(0)$. Consider the norm of the gradient of the loss with respect to $\vb_k$,
\begin{align*}
\|\nabla_{\vb_k}L(s) \|_2= \bigg \|\frac{1}{\sqrt{m}}\sum_{i =1 }^{n}(f_i(s) - y_i) \frac{c_kg_k(s)}{\|\vb_k(s)\|_2}\ind_{ik}(s)\xb_i^{\vb_k(s)^\perp} \bigg\|_2.
\end{align*} Since $\|\vb_k(s)\|_2\ge  \|\vb_k(0)\|_2\ge  \alpha (\delta / m)^{1 / d}$ with probability $1- \delta$ over the initialization, applying Cauchy Schwartz's inequality gives
\begin{align}\label{eq:grad_v_bound}
 \|\nabla_{\vb_k}L(s) \|_2 \le \frac{(1 + R_g( m / \delta) ^{1 / d})\sqrt{n} \|\fb (s) - \yb \|_2}{\alpha \sqrt{m}}.
\end{align}
By the assumption on $\|\fb(s) - \yb\|_2$ this gives
\begin{align*}
\|\nabla_{\vb_k}L(s)\|_2\le \frac{2\sqrt{n} (1 - \frac{\eta \omega}{2})^{s/2}\|\fb (0) - \yb \|_2}{\alpha \sqrt{m}}.
\end{align*}
Hence bounding the parameter trajectory by the sum of the gradient updates:

\begin{align*}
\|\vb_k(K+1) - \vb_k(0)\|_2\le \eta \sum_{s = 0}^{K} \|\nabla_{\vb_k}L(s) \|_2  \le  \frac{2\sqrt{n} \|\fb (0) - \yb\|_2}{\alpha \sqrt{m}}\sum_{s=1}^{K} \bigg(1 - \frac{\eta \omega}{2} \bigg)^{s/2}
\end{align*}
yields a geometric series. Now the series is bounded as
\begin{align*}
\frac{1}{1 -\sqrt{1 - \frac{\eta \omega}{2}}} \le \frac{4}{\eta \omega},
\end{align*}
which gives
\begin{align*}
\|\vb_k(K+1) - \vb_k(0) \|_2\le \frac{8\sqrt{n}\|\fb (0) -  \yb\|_2}{\alpha \sqrt{m} \omega}.
\end{align*}
\end{proofLemma}

\raggedbottom
\end{document}